\RequirePackage{letltxmacro}
\documentclass[pdflatex,sn-mathphys]{sn-jnl}

\jyear{2021}%

\usepackage[T5]{fontenc}
\usepackage{tabulary}
\usepackage{latexsym}
\usepackage{float}
\usepackage{makecell}
\usepackage{array,booktabs,arydshln,xcolor}
\usepackage{subcaption,booktabs}
\usepackage{microtype}
\usepackage{multirow}
\usepackage{multicol}
\usepackage{threeparttable} 
\usepackage{amsmath}
\usepackage{tablefootnote}
\usepackage{pifont}
\usepackage{float}
\usepackage{caption}
\usepackage{placeins}
\usepackage{array}
\usepackage{textcomp}
\usepackage{lmodern}
\usepackage{graphicx}
\usepackage{soul,color}
\usepackage{xurl}
\usepackage{amssymb}
\usepackage{pifont}
\usepackage[normalem]{ulem}
\usepackage{xcolor}

\theoremstyle{thmstyleone}%
\theoremstyle{thmstyletwo}%

\theoremstyle{thmstylethree}%
\newtheorem{definition}{Definition}%
\begin{document}

\title[Weakly Supervised Data Labeling Framework for Lexical Normalization]{A Weakly Supervised Data Labeling Framework for Machine Lexical Normalization in Vietnamese Social Media}



\author[1,2]{\fnm{Dung} \spfx{Ha} \sur{Nguyen}}\email{20520165@gm.uit.edu.vn}

\author[1,2]{\fnm{Anh} \spfx{Thi Hoang} \sur{Nguyen}}\email{20520134@gm.uit.edu.vn}

\author*[1,2]{\fnm{Kiet} \sur{Van Nguyen}}\email{kietnv@uit.edu.vn}

\affil[1]{\orgname{University of Information Technology}, \orgaddress{ \city{Ho Chi Minh City}, \country{Vietnam}}}

\affil[2]{\orgname{Vietnam National University}, \orgaddress{ \city{Ho Chi Minh City}, \country{Vietnam}}}

\abstract{This study introduces an innovative automatic labeling framework to address the challenges of lexical normalization in social media texts for low-resource languages like Vietnamese. Social media data is rich and diverse, but the evolving and varied language used in these contexts makes manual labeling labor-intensive and expensive. To tackle these issues, we propose a framework that integrates semi-supervised learning with weak supervision techniques. This approach enhances the quality of training dataset and expands its size while minimizing manual labeling efforts. Our framework automatically labels raw data, converting non-standard vocabulary into standardized forms, thereby improving the accuracy and consistency of the training data. Experimental results demonstrate the effectiveness of our weak supervision framework in normalizing Vietnamese text, especially when utilizing Pre-trained Language Models. The proposed framework achieves an impressive F1-score of 82.72\% and maintains vocabulary integrity with an accuracy of up to 99.22\%. Additionally, it effectively handles undiacritized text under various conditions. This framework significantly enhances natural language normalization quality and improves the accuracy of various NLP tasks, leading to an average accuracy increase of 1-3\%.}


\maketitle
\section{Introduction} \label{section-1}

The explosive growth of social media has transformed how we communicate, generating an immense volume of informal, dynamic text. Platforms like Facebook, X, and Instagram contribute billions of posts daily, each encapsulating diverse linguistic, cultural, and personal expressions. According to LocaliQ\footnote{\url{https://localiq.com/blog/what-happens-in-an-internet-minute/}}, every minute in 2024 sees about 510,000 posts, 4 million Facebook post likes, 360,000 tweets on X, 694,000 video reels shared on Instagram, 41.6 million WhatsApp messages, and over 6 million Google searches. This abundant yet chaotic data presents a significant challenge for Natural Language Processing (NLP), especially in low-resource languages such as Vietnamese.


For low-resource languages, the problem is compounded by the lack of large, well-annotated datasets essential for training robust NLP models. Vietnamese social media language, in particular, is marked by several complexities:

\begin{enumerate}
    \item \textit{Linguistic Diversity}: Users from different regions bring varying vocabularies and grammatical structures, complicating the creation of standardized linguistic norms.
    \item \textit{Spelling Errors and Typos}: Informal communication on social media often leads to a high frequency of spelling errors and typos, introducing additional noise for NLP models to handle.
    \item \textit{Emergence of Neologisms and Slang}: Social media often features new terms, creative expressions, and slang, which are typically absent from traditional dictionaries and standardized grammar.
    \item \textit{Rapid Evolution}: The language used on social media evolves quickly, with trends and new words appearing and disappearing rapidly, posing a challenge for consistent vocabulary tracking and normalization.
    \item \textit{Foreign Language Influence}: Frequent incorporation of foreign terms, particularly English, further complicates efforts to standardize Vietnamese vocabulary.
\end{enumerate}

This study focuses on addressing two main challenges: spelling errors and typos, and the emergence of neologisms. Spelling and typing inaccuracies arise from users' unfamiliarity with the language or lack of meticulousness, exacerbated by the omission of diacritics which leads to ambiguity and comprehension difficulties. Neologisms emerge as users prioritize efficiency in messaging, resorting to abbreviations, teencode, and slang, particularly among younger demographics.

These challenges underscore the critical need for lexical normalization, which is transforming non-standard words (NSWs) into their standard forms, in social media. While normalization may reduce some unique features of social media language \cite{nguyen-etal-2021-learning}, it has been shown to improve performance in numerous downstream NLP tasks, such as named entity recognition \cite{plank-etal-2020-dan}, POS tagging \cite{Zupan_Ljubešić_Erjavec_2019}, dependency and constituency parsing \cite{van-der-goot-van-noord-2017-parser}, sentiment analysis \cite{https://doi.org/10.25932/publishup-43742}, and machine translation \cite{bhat-etal-2018-universal}. Conventional NLP tools, primarily designed for standardized text, often struggle with the informal, ever-changing nature of social media language, making effective normalization essential.

However, the traditional approach to building datasets for lexical normalization through manual annotation is impractical for low-resource languages due to the significant labor, expertise, and cost involved. To address these challenges, we propose a weakly supervised data labeling framework for lexical normalization of Vietnamese social media, inspired by the approach in \cite{karamanolakis-etal-2021-self}. This framework leverages a combination of weakly supervised and semi-supervised learning methods to automatically label and expand training datasets, thereby reducing the need for manual annotation and lowering associated costs.

Our research has three primary scientific contributions:
\begin{itemize}
   \item Firstly, we create a weakly supervised labeling framework that utilizes both weak supervision and semi-supervised learning to efficiently generate and expand high-quality training data for machine lexical normalization in social media, marking a novel approach in the field.
   \item Next, we optimize existing pre-trained models to effectively normalize Vietnamese social media text, transforming NSWs into their standard equivalents, establishing a pioneering contribution. This includes the development of a specialized tokenizer to maintain alignment between standard and non-standard tokens, improvement of pre-trained language models (ViSoBERT, PhoBERT, and BARTpho) to dynamically manage token adjustments during normalization, and refinement of the overall architecture using an innovative strategy tailored to handle noise effectively.
   \item Finally, we also evaluate the impact of lexical normalization on various real NLP applications, including hate speech detection, emotion recognition, hate speech span detection, spam review detection, and aspect-based sentiment analysis.
\end{itemize}

Our framework is the first to propose a weakly supervised approach for lexical normalization in low-resource languages. Focused initially on Vietnamese, it offers a scalable solution adaptable to similar language challenges. By generating large, high-quality datasets cost-effectively, it enhances NLP performance in dynamic social media contexts, bridging the gap between unstructured data and NLP model requirements.

The paper is structured as follows: Section \ref{section-2} provides background on lexical normalization and data labeling, along with related work. Section \ref{section-3} introduces the experimental datasets. Section \ref{section-4} details our proposed data labeling framework for lexical normalization. Section \ref{section-5} presents experimental results. Finally, Section \ref{section-6} concludes with a summary and future research directions.

\section{Fundations of Lexical Normalization and Data Labeling} \label{section-2}
\subsection{Lexical Normalization} \label{subsec:2-1}
The job of lexical normalization, as defined by van der Goot et al. \cite{van-der-goot-etal-2021-multilexnorm}, is limited to social media data and can be represented as follows in this paper: 
\begin{definition}
Lexical normalization is the task of transforming an utterance into its standard form, word by word, including both one-to-many (1-n) and many-to-one (n-1) replacements.
\end{definition}
It should be noted that lexical normalization relies on a certain degree of correspondence between the words of target sentence and the terms of source sentence. Therefore, we did not add, remove, or rearrange any words in the sentence while it is being standardized.

\begin{table}[]
\centering
\resizebox{\columnwidth}{!}{%
\begin{tabular}{|c|l|l|l|}
\hline
\textbf{Mapping Type} &
  \multicolumn{1}{c|}{\textbf{1-1}} &
  \multicolumn{1}{c|}{\textbf{1-n}} &
  \multicolumn{1}{c|}{\textbf{n-1}} \\ \hline
\textbf{Source} & hôm nay \underline{t} \underline{ko} đi học      & \underline{kt} dạo này bất ổn ghê      & thằng \underline{ch} \underline{ó} \underline{chế} \underline{t} \\ \hline
\textbf{Target} & hôm nay \underline{tao} \underline{không} đi học & \underline{kinh tế} dạo này bất ổn ghê & thằng \underline{chó} \underline{chết}   \\ \hline
\textbf{English} &
  today i don't go to school &
  \begin{tabular}[c]{@{}l@{}}the economy is extremely \\ unstable these days\end{tabular} &
  son of a bitch \\ \hline
\end{tabular}%
}
\caption{Examples illustrating sentences containing NSW and their standard forms.}
\label{lexical_example}
\end{table}

Due to the dynamic nature of social network language, characterized by rapid evolution and regional/demographic variation, establishing conventions for standard and non-standard forms is crucial. This study identifies and categorizes the following types of NSWs within our training set.
\begin{itemize}
    \item \textbf{Abbreviations}: Shortenings intentionally created by users to expedite communication on social networks. For example:  ``mng'' $\rightarrow$ ``mọi người'' (everyone) and ``ko'' $\rightarrow$ ``không'' (no).
    \item \textbf{Spelling errors}: Mistakes arising from user knowledge gaps or typos. For example: ``ddungss'' $\rightarrow$ ``đúng'' (right) and ``suy nghỉ'' $\rightarrow$ ``suy nghĩ'' (thought).
    \item \textbf{Teencode}: A unique form of language, including special symbols and symbols to secretly convey messages, is the way the young generation ``encodes'' language in text communication. For example: ``chầm zn'' $\rightarrow$ ``trầm cảm'' (depression), ``trmúa hmề'' $\rightarrow$ ``chúa hề'' (clown).
    \item \textbf{Unmarks}: Omissions of diacritics within sentences, potentially due to typos, time constraints, or mitigating the impact of sensitive words. For example: ``goi st5k da het thoi gian su dung'' $\rightarrow$ ``gói st5k đã hết thời gian sử dụng'' (``st5k package has expired''), ``du di me'' $\rightarrow$ ``đụ đĩ mẹ'' (motherfucker).
\end{itemize}
Furthermore, the research scope encompasses non-standard elements beyond the aforementioned categories, including emoji, emoticons, @username, hashtags, phone numbers, URLs, and email addresses, etc.

\subsection{Existing Lexical Normalization Methods}
While lexical normalization remains a nascent field of research,  preliminary studies have already demonstrated its significant potential. The W-NUT workshop established a pioneering shared task on lexical normalization, focusing on User-Generated Content (UGC) data from English tweets \cite{baldwin-etal-2015-shared}. Subsequent research explored various machine learning approaches for this task. For instance, one study employed a Random Forest Classifier to predict the appropriate normalized form of words, leveraging features like support, confidence, string similarity, and POS tags \cite{jin-2015-ncsu}. Other researchers investigated the application of Conditional Random Fields (CRF) models \cite{akhtar-etal-2015-iitp-hybrid, supranovich-patsepnia-2015-ihs} and recurrent neural networks (RNN) models \cite{min-mott-2015-ncsu, wagner-foster-2015-dcu} to this problem.

MoNoise \cite{van-der-goot-2019-monoise} stands as a prominent technology in vocabulary normalization. This model, initially designed for English text, leverages spelling correction and word embeddings to generate candidate normalized forms, which are then ranked using a Feature-Based Random Forest Classifier. Later advancements extended the capabilities of MoNoise to standardize multilingual vocabularies, including Dutch, Spanish, Turkish, Slovenian, Croatian, and Serbian.

Pre-trained language models (PLMs) have emerged as an alternative approach for textual standardization. Research by \cite{muller-etal-2019-enhancing} demonstrates that the BERT model \cite{devlin-etal-2019-bert} can effectively tackle this task by framing it as a token prediction problem. This approach proves particularly efficient when training data is limited.  Furthermore, \cite{bucur-etal-2021-sequence} explores the application of the pre-trained mBART model \cite{tang-etal-2021-multilingual} as an encoder for automatic de-noising, essentially translating ``bad English'' sentences into grammatically correct ones. This method offers the advantage of considering entire sentences during normalization and can be readily extended to additional languages without significant computational overhead.

Compared to well-resourced languages like English, research on textual standardization in Vietnamese is in its early stages. However, recent studies demonstrate promising progress. In 2016, a method for standardizing Vietnamese Twitter data employed the Dice coefficient or n-gram for spelling correction, followed by SVM classification with various feature types \cite{Nguyen2016}. Subsequent research in 2019 focused on speech-to-text data standardization using a combined approach \cite{10.1145/3342827.3342851}. Random Forest initially classified NSWs, with subsequent steps tailored to the NSW type. Numerical text was expanded using defined rules, letter types were standardized with a deep learning sequence-to-sequence model, and pronunciation was derived for other NSWs. In 2022, another study addressed text standardization for Text-to-Speech (TTS) systems using news clippings \cite{9953791}. This work employed model-based tagging tools (CRF, BiLSTM-CNN-CRF, and BERT-BiGRU-CRF) to identify NSWs, followed by rule-based normalization (Forward Lexicon-Based Maximum Matching algorithm) based on the NSW type.

The most recent advancements tackle Vietnamese lexical normalization directly, particularly for social media language. Research in 2023 by \cite{hsd_lexnorm} proposes a Sequence-to-Sequence model for lexical normalization to enhance Hate Speech Detection accuracy.  Complementing this approach, \cite{nguyen-etal-2024-vilexnorm} established ViLexNorm, a large Vietnamese social network text standardization dataset. ViLexNorm has been explored on modern approaches, such as transformer-based models, for the standardization process.

\subsection{Data Labeling}
This section provides a foundational overview of data labeling, a critical step in supervised machine learning tasks.


Data labeling is the cornerstone of supervised machine learning, where models learn to map inputs to desired outputs based on labeled examples. This process involves assigning labels or additional information to raw data (text, images, audio, video) to create training datasets.
\begin{definition}
    For lexical normalization, data labeling is treated as a token-level sequence labeling problem, where an input sentence $x=x_1^n=(x_{1}, x_{2}, \dotsm, x_{n})$ consists of $n$ tokens, and each token $x_{i}$ has a corresponding label $y_{i}$ in the gold label sequence $y=y_1^n=(y_{1}, y_{2}, \dotsm, y_{n})$. The labeled dataset $D_L$ is a set of pairs of sentences and their corresponding gold label sequences, represented as $D_L =\lbrace (x, y) \rbrace$, where each pair $(x, y)$ consists of a sequence of tokens $x$ and its corresponding sequence of labels $y$.
\end{definition}
In NLP tasks, labeling can encompass tasks like POS tagging (identifying nouns, verbs, etc.), sentiment analysis (positive, negative), or topic categorization. There are three primary approaches to data labeling:
\begin{itemize}
    \item \textit{Manual Labeling}: This method relies on human experts to meticulously assign labels to data points. It necessitates a deep understanding of the labeling context and goals, alongside precision and consistency. While manual labeling guarantees high-quality and accurate labels, which are crucial for robust machine learning models, it can be time-consuming and resource-intensive, especially for large datasets. Imagine image recognition tasks where specialists must manually label objects like cars, pedestrians, or traffic signs in thousands of images.

    \item \textit{Automatic Labeling}: This approach leverages algorithms or pre-trained machine learning models to automate the labeling process, significantly reducing time and effort, particularly with extensive datasets. Popular techniques include pre-trained models, clustering algorithms, and rule-based methods. While automatic labeling may not achieve the same accuracy as manual labeling, it serves as a valuable tool for creating preliminary datasets or streamlining the manual process by reducing the initial workload.

    \item \textit{Semi-Automatic Labeling}: This method strategically combines manual and automatic approaches to capitalize on their respective strengths. Automatic algorithms provide initial labels, which are then meticulously reviewed and refined by human experts to ensure accuracy. This approach fosters efficiency and speed in the labeling process while maintaining the quality of labeled data. It is often utilized in large projects where manual labeling becomes infeasible due to data volume, but high accuracy remains paramount.
\end{itemize}
The quality of data labels directly impacts the performance of a machine learning model. Inaccurate or inconsistent labels can lead to suboptimal or even erroneous models. Therefore, it is critical to evaluate and control labeling quality. Common methods include utilizing multiple annotators and analyzing their agreement (inter-annotator agreement), randomly sampling and checking labeled data, and employing dedicated test sets. Additionally, automatic quality check methods can be implemented.

\subsection{Existing Data Labeling Approaches}
Deep learning has revolutionized various tasks and domains due to its exceptional performance. However, this success hinges on training complex architectures with labeled data. Manually collecting, managing, and labeling this data is a significant bottleneck, prompting renewed interest in classical approaches that address label scarcity.

Active Learning (AL) \cite{10.1145/3472291} tackles this by selecting the most informative data points for labeling, optimizing label allocation. Semi-Supervised Learning (SSL) \cite{10.5555/3495724.3496249} leverages unlabeled data to improve model performance. Transfer Learning (TL) \cite{10.1145/3400066} pre-trains a model on a similar domain to enhance performance on a new one. While these techniques are valuable, they still require a foundation of labeled data, limiting their ability to fully address label scarcity.

To alleviate labeling burden, researchers have explored alternative, less expensive sources of labels. Distant supervision utilizes external knowledge sources to generate noisy labels \cite{hoffmann-etal-2011-knowledge}. Crowdsourcing \cite{6113213}, heuristic rules \cite{Awasthi2020Learning}, and feature annotation \cite{JMLR:v11:mann10a} are other approaches. A key question lies in combining these methods effectively and in a principled manner.

Programmatic Weak Supervision (PWS) offers a compelling answer \cite{10.14778/3157794.3157797}. In PWS, users define weak supervision sources (e.g., heuristics, knowledge bases, pre-trained models) as Labeling Functions (LFs). These user-defined programs provide labels for specific data subsets, collectively creating a vast training set. However, LFs can introduce noise at varying rates and may generate conflicting labels for some data points. To address these issues, researchers developed label models that synthesize labels generated by LFs \cite{10.5555/3157382.3157497}. These labels are then used to train the final model. Recent work explores combining label models and the final model into joint models \cite{ren-etal-2020-denoising}.

PWS demonstrates flexibility by integrating with other machine learning methods. It can enhance AL by querying labels that best align with existing labels based on LFs \cite{Mallinar_Shah_Ho_Ugrani_Gupta_2020}. Conversely, AL can assist PWS by directing expert labeling towards data points that challenge the label model. PWS can also be synergistically combined with TL. While TL reduces the need for pre-labeled data, it doesn't eliminate it entirely. PWS can provide these necessary labels. Modern PWS methods often leverage pre-trained models fine-tuned with labels generated by the label model \cite{zhang2021wrench}. Additionally, SSL's concept of leveraging unlabeled data has been applied to PWS \cite{karamanolakis-etal-2021-self}, where self-training is used to label unlabeled data and augment the training set. In essence, SSL and PWS combine clean and weak labels to boost model performance.
\subsection{A Weakly Supervised Approach to Data Labeling for Lexical Normalization in Vietnamese Social Media}
While lexical normalization has received attention in other languages, research specifically focused on Vietnamese social media text is scarce. Existing studies have limited exploration of PLMs for this task and haven't extensively addressed data from contemporary online platforms like social networks. To address these gaps, we propose a direct text normalization approach, bypassing an error detection step, that leverages datasets extracted from popular Vietnamese social media platforms such as Facebook, TikTok, and YouTube. This data is rich in NSWs, and the proposed model aims to both generalize and map these NSWs to their standard forms.

Inspired by prior research \cite{nguyen-etal-2023-visobert, nguyen-tuan-nguyen-2020-phobert, DBLP:journals/corr/abs-2109-09701}, we propose applying and fine-tuning prominent Vietnamese PLMs like PhoBERT, BARTPho, and ViSoBERT for Vietnamese textual standardization. This approach capitalizes on the capabilities of modern language models, potentially improving the accuracy and efficiency of normalization specifically within the context of Vietnamese social media data.

In the lexical normalization task, raw data consists of sentences containing one or more NSWs. Each NSW requires labeling with its corresponding standard form. The objective of model is to transform sentences containing NSWs into sentences where these NSWs have been normalized. Existing textual standardization studies have relied on manually pre-labeled datasets, often limited in size (e.g., 2,000 sentences). Our work proposes a novel approach that combines Pseudo-Labeling with Self-Training (PWS-ST) within the SSL framework, inspired by \cite{karamanolakis-etal-2021-self}. This approach tackles data scarcity by automatically expanding the training set size. The self-training algorithm trains a prediction model on pre-labeled data to generate pseudo-labels for unlabeled data. PWS utilizes LFs to assign weak labels to unlabeled data.  Furthermore, we incorporate a neural network to assess the quality of these weak labels, ensuring only high-confidence labels are added to the training data for model learning.

Ultimately, our work aims to develop an automatic labeling framework for text normalization, a crucial task in NLP, particularly when dealing with social media data characterized by frequent use of non-standard language. Unlike sequence classification tasks where the model predicts a single label for the entire input, text normalization requires sequence labeling, assigning a label (standard form) to each individual token within the input sentence. This necessitates the ability of model to not only identify NSWs but also accurately map each NSW to its corresponding standard word. This complex process demands a high degree of sophistication in handling sentence context and semantics.
\section{Datasets and Data Pre-processing} \label{section-3}
\subsection{Datasets}
\textbf{ViLexNorm} \cite{nguyen-etal-2024-vilexnorm}, specifically designed for vocabulary standardization, is collected from two prominent Vietnamese social media platforms: Facebook and TikTok. It comprises 10,467 comment pairs, each containing an original sentence and its corresponding post-normalized version.

This research leverages the ViLexNorm dataset for two key purposes:

(1) \textbf{Fine-tuning the pre-trained language model}: We employ the ViLexNorm dataset to fine-tune a pre-trained language model for the task of Vietnamese vocabulary standardization. This process enhances the model's ability to identify and normalize informal or dialectal variations commonly found in Vietnamese social media text.

(2) \textbf{Pre-labeled data for labeling framework construction}: The ViLexNorm dataset serves as a pre-labeled dataset (denoted as $D_{L}$) within the framework we construct for model training. This dataset provides valuable seed labels to guide the learning process.

To comprehensively assess the generalizability of model across various Vietnamese social media text analysis tasks, we incorporate five publicly available datasets. A detailed description of each dataset is provided in Table \ref{tab:datasets}.

\begin{itemize}
    \item \textbf{ViHSD} \cite{10.1007/978-3-030-79457-6_35}: This dataset encompasses user comments on various topics including entertainment, celebrities, social and political issues. Extracted from Facebook and YouTube, ViHSD contains a total of 33,400 comments labeled into three categories: CLEAN, OFFENSIVE, and HATE. This dataset serves the task of hate speech detection.
    \item \textbf{UIT-VSMEC} \cite{10.1007/978-981-15-6168-9_27}: Designed for emotion recognition, the UIT-VSMEC dataset includes comments extracted from public Facebook posts. It comprises 6,927 comments, each assigned one of six emotion labels: Enjoyment, Sadness, Anger, Fear, Disgust, Surprise, and Other.
    \item \textbf{ViHOS} \cite{hoang-etal-2023-vihos}: This dataset focuses on Vietnamese hate and offensive span detection. ViHOS includes 11,056 comments with labeled spans of offensive or hateful content.
    \item \textbf{ViSpamReviews} \cite{10.1007/978-3-031-21743-2_48}: Collected from leading Vietnamese online shopping platforms, ViSpamReviews serves the tasks of spam detection and classification. This dataset includes 19,870 rows containing customer reviews and product/service comments.
    \item \textbf{UIT-ViSFD} \cite{10.1007/978-3-030-82147-0_53}: Announced for aspect-based sentiment analysis (ABSA) and sentiment classification, UIT-ViSFD consists of customer reviews on e-commerce platforms for 10 popular smartphone brands in Vietnam. The dataset includes 11,122 comments labeled with three sentiment labels (Positive, Negative, and Neutral) and 10 aspect labels related to smartphone features (e.g., SCREEN, CAMERA, and BATTERY).
\end{itemize}

\begin{table}[]
\centering
\resizebox{\columnwidth}{!}{%
\begin{tabular}{ccccc}
\hline
\textbf{Dataset} &
  \textbf{Instances} &
  \textbf{Features} &
  \textbf{Domain} &
  \textbf{Classes} \\ \hline
ViLexNorm     & 10,467 & 2 & Lexical Normalization              & -   \\
ViHSD         & 33,400 & 2 & Hate Speech Detection              & 3   \\
UIT-VSMEC     & 6,927  & 2 & Emotion Recognition                & 7   \\
ViHOS         & 11,056 & 4 & Hate and Offensive Spans Detection & -   \\
ViSpamReviews & 19,870 & 4 & Spam Detection/Classification      & 2/4 \\
UIT-ViSFD &
  11,122 &
  5 &
  \begin{tabular}[c]{@{}c@{}}Aspect-Based Sentiment Analysis/\\ Sentiment Classification\end{tabular} &
  10/3 \\ \hline
\end{tabular}%
}
\caption{Dataset description and statistical source.}
\label{tab:datasets}
\end{table}


\subsection{Data Preprocessing}
\begin{figure}[ht]
    \centering
    \includegraphics[width=\linewidth]{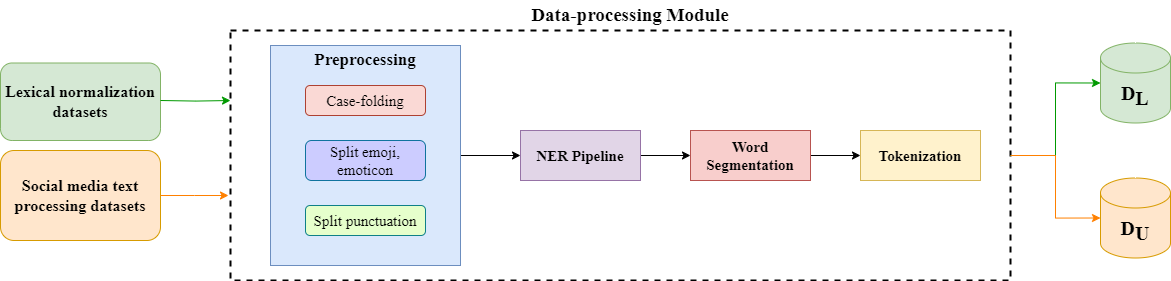}
    \caption{Preprocessing and Data Generation Workflow for the Framework.}
    \label{fig:data_preprocessing}
\end{figure}
This section details the data preparation process employed in our research. We leverage two primary data sources:
\begin{itemize}
    \item Lexical normalization datasets ($D_{L}$): This dataset serves as the foundation for generating labeled data ($D_{L}$).  We utilize the normalized labels within ViLexNorm to guide the creation of $D_{L}$.
    \item Social media text processing datasets ($D_{U}$): We exploit five publicly available datasets (ViHSD, UIT-VSMEC, ViHOS, ViSpamReviews, and UIT-ViSFD) to extract comments and sentences for the creation of unlabeled data ($D_{U}$). These datasets provide a rich source of real-world Vietnamese social media text.
\end{itemize}

The data preparation process incorporates several key steps as outlined in the provided module:
\subsubsection{Basic Preprocessing}
Most of the utilized datasets have undergone basic preprocessing procedures during their initial publication. Therefore, our team focused on addressing task-specific issues related to text normalization.
\begin{itemize}
    \item \textbf{Case-folding}: This step ensures consistency by converting all text to lowercase. This mitigates variations arising from capitalization differences. For example: ``Hôm nay T hk đi học, t đi chơi'' $\rightarrow$  ``hôm nay t hk đi học, t đi chơi'' (\textbf{English}: Today I don't go to school, I go out)
      
    \item \textbf{Emoticon/Emoji Separation}: Emoticons and emojis are prevalent in social media communication. To prevent these elements from introducing unwanted variations when attached to text, we separate them using spaces. For example:
    
    \begin{itemize}
        \item ``Sao m nói dị hả\includegraphics[height=1em]{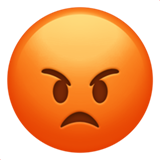}\includegraphics[height=1em]{Fig/pouting-face.png}'' $\rightarrow$  ``sao m nói dị hả \includegraphics[height=1em]{Fig/pouting-face.png}\includegraphics[height=1em]{Fig/pouting-face.png}'' (\textbf{English}: Why do you say that \includegraphics[height=1em]{Fig/pouting-face.png}\includegraphics[height=1em]{Fig/pouting-face.png})
        \item ``mắc cừi quá=))'' $\rightarrow$ ``mắc cừi quá =))'' (\textbf{English}: So funny =)))
    \end{itemize}
    
    \item \textbf{Punctuation Separation}: Similar to emoticons/emoticons, punctuation marks can sometimes be irregularly attached to text, leading to lexical variations. We address this by separating punctuation marks with spaces. For example: ``B ơi!! coi cái này nè….'' $\rightarrow$ ``B ơi !! coi cái này nè ….'' (\textbf{English}: Hey bro !! Look at this ….)
\end{itemize}
\subsubsection{Named Entity Recognition (NER) Pipeline}
Protecting user privacy is essential in social media data analysis. Social network texts can contain sensitive information like usernames and email addresses. To ensure anonymity, we use Named Entity Recognition (NER) to identify and mask such details.

Our method applies the NER Pipeline by NlpHUST\footnote{\url{https://huggingface.co/NlpHUST/ner-vietnamese-electra-base}} to detect person names, replacing them with placeholders like ``@username'' This approach safeguards user identities while adhering to data protection regulations. For example, ``Phương Nguyễn ê ik chỗ này điiii'' is anonymized to ``@username ê ik chỗ này điiii'' (\textbf{English}: @username hey let's go to this place).

\subsubsection{Word Segmentation}
Word segmentation, a crucial step in NLP, involves dividing a continuous character stream into individual words. While languages like English rely on spaces for word boundaries, languages such as Vietnamese, Chinese, and Japanese lack these clear delimiters, making segmentation more complex.

The performance of word segmentation significantly impacts downstream tasks.  In this work, we leverage the VnCoreNLP\footnote{\url{https://github.com/vncorenlp/VnCoreNLP}} tool, a popular NLP suite developed by the Institute of Information and Communications Technology (ICT) at Hanoi University of Science and Technology. VnCoreNLP utilizes a machine learning model trained on extensive Vietnamese corpora, enabling effective and accurate word segmentation. This tool facilitates efficient processing of large text volumes with high accuracy. For example: ``Hạnh Nguyễn con mẹ nó đoạn hôm qua t làm rồi nhưng mà nay lỗi'' $\rightarrow$ ``@username con\_mẹ nó đoạn hôm\_qua t làm rồi nhưng\_mà nay lỗi'' (\textbf{Englsih}: @username damn it, the part I did yesterday is now broken).

\subsubsection{Tokenization}
Tokenization, a fundamental step in NLP, follows word segmentation and involves splitting text into even smaller units called tokens. These tokens can encompass individual words, punctuation marks, or even special characters.

Although spaces effectively separate words in English and other Latin languages, Vietnamese requires additional considerations after segmentation. In our approach, we leverage the results of the word segmentation step to create tokens by utilizing spaces as delimiters. This approach ensures accurate tokenization for Vietnamese text analysis. For example: ``Phương Ng mốt có bồ rồi đăng đỡ si nghĩ '' $\rightarrow$ [`@username’, `mốt', `có', `bồ', `rồi', `đăng', `đỡ', `si nghĩ']

\subsection{Dataset Description}
After processing, the labeled dataset ($D_{L}$) consists of 10,463 entries, each with corresponding labels as exemplified in Table 3.8. The data is divided into three sets: training, development, and test, with a ratio of 8:1:1. The structure of the dataset is illustrated in Table \ref{tab:data_description}.

\begin{figure}
    \centering
    \includegraphics[width=0.6\linewidth]{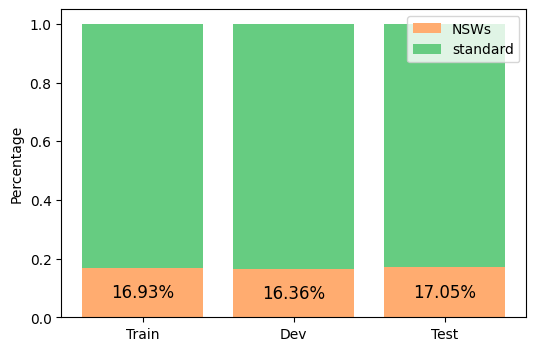}
    \caption{NSWs Proportion in ViLexNorm Train, Dev, and Test Dataset.}
    \label{fig:nsw-rate}
\end{figure}

\begin{table}[ht]
\label{tbl:att}
\centering
\resizebox{\columnwidth}{!}{%
\begin{tabular}{llc}
\hline
\multicolumn{1}{c}{\textbf{Attribute}} & \multicolumn{1}{c}{\textbf{Description}} & \textbf{Data Type} \\ \hline
original   & Original comment sentence, unstandardized        & String \\
normalized & Normalized comment sentence with assigned labels & String \\
input      & Sequence of tokens from the original sentence    & Object \\
output     & Sequence of tokens from the normalized sentence  & Object \\ \hline
\end{tabular}%
}
\caption{Attributes of the Labeled Dataset.}
\label{tab:data_description}
\end{table}
Figure \ref{fig:nsw-rate} illustrates the proportion of vocabulary that requires normalization in the training, development, and test sets. In all three datasets, the proportion of words needing normalization is approximately equal, around 17\%. 

\begin{figure}
    \centering
    \includegraphics[width=\linewidth]{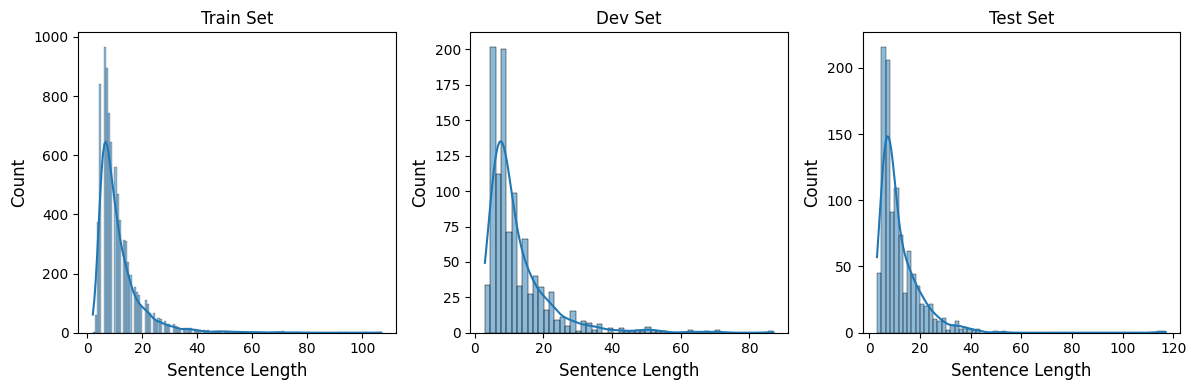}
    \caption{The Distribution of Sentence Length of ViLexNorm Train, Dev, and Test Dataset.}
    \label{fig:lendist}
\end{figure}

Figure \ref{fig:lendist} shows the distribution of sentence lengths across the three datasets. In all sets—training, development, and test—the sentence lengths predominantly range from 1 to 40 words. The longest sentence in the ViLexNorm dataset contains nearly 120 words.

The unlabeled dataset ($D_{U}$) is processed and compiled from five datasets, retaining the original dataset names and the names of the extracted subsets. The final dataset comprises a total of 121,087 entries with six attributes described in Table 

\begin{table}[ht]
\centering
\resizebox{\columnwidth}{!}{%
\begin{tabular}{llc}
\hline
\multicolumn{1}{c}{\textbf{Attribute}} &
  \multicolumn{1}{c}{\textbf{Description}} &
  \textbf{Data Type} \\ \hline
dataset &
  \begin{tabular}[c]{@{}l@{}}Name of the datasets \\ (ViHSD, UIT-VSMEC, ViHOS, ViSpamReviews, UIT-ViSFD)\end{tabular} &
  STRING \\
type      & Extracted subset of the dataset (train, dev, test) & STRING \\
sent\_idx & Original index from the extracted dataset          & INT    \\
idx       & Index after sentence segmentation from sent\_idx   & INT    \\
original  & Original comment sentence, unstandardized          & STRING \\
input     & Sequence of tokens from the original sentence      & OBJECT \\ \hline
\end{tabular}%
}
\caption{Attributes of the Unlabeled Dataset.}
\label{tab:d_U_description}
\end{table}
\section{Our Proposed Framework} 
\label{section-4}
\subsection{Overall Framework Architecture} 
\label{proposed_framework}

\begin{figure}[ht]
    \centering
    \includegraphics[width=0.8\linewidth]{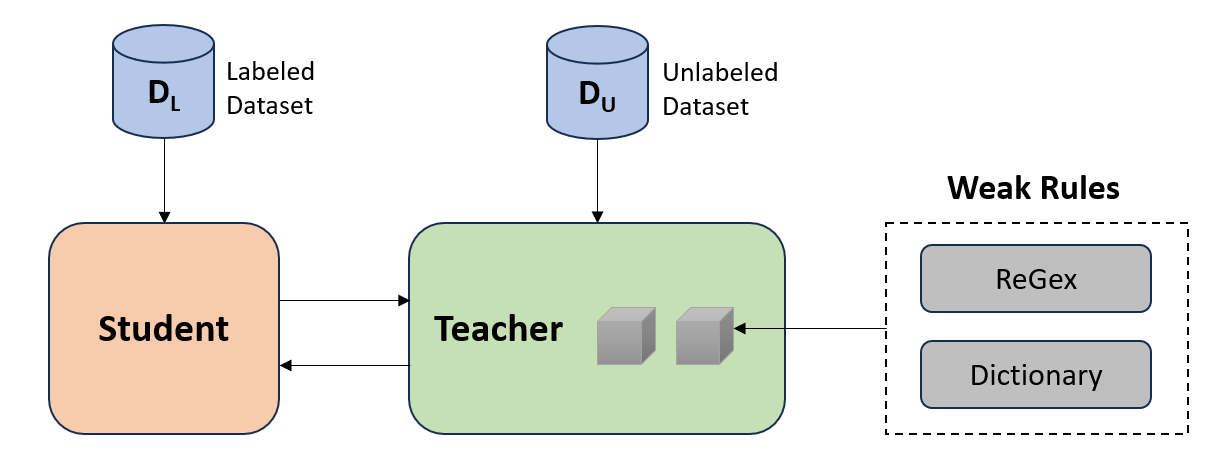}
    \caption{Main Components in the Architecture of the Proposed Framework.}
    \label{fig:framework}
\end{figure}

The proposed framework draws inspiration from the ASTRA labeling framework \cite{karamanolakis-etal-2021-self} and is designed to harness both labeled and unlabeled data for lexical normalization tasks. As shown in Figure \ref{fig:framework}, the architecture employs a weak supervision paradigm, utilizing rules as weak supervision sources to generate weak labels for unlabeled data. At its core, the framework integrates a Student model, trained on labeled datasets, capable of producing pseudo-labels for unlabeled text. Complementing this is the Teacher model, which employs a Rule Attention Network (RAN) to integrate pseudo-labels from the Student model with weak labels derived from various weak sources, thus generating final predicted labels for unlabeled data. The detailed architecture and training procedure of the labeling framework are elaborated in subsequent sections.

\subsection{Student Base Model} \label{student}

The Student model serves as the initial and crucial component of the proposed framework. Initially, the Student model is trained on a labeled dataset $D_L$. This trained model is subsequently employed to generate pseudo-labels for samples in the unlabeled dataset $D_U$. In traditional self-training methods, these pseudo-labeled samples are directly incorporated into the training data, and the model undergoes iterative training. However, the proposed framework enhances this self-training process by integrating weak labels from the Teacher model (detailed in Section \ref{teacher}).

In this study, the Student model utilized comprises pre-trained language models tailored for Vietnamese, addressing the specific challenge of lexical normalization. Specifically, we experiment with three models: ViSoBERT \cite{nguyen-etal-2023-visobert}, PhoBERT \cite{nguyen-tuan-nguyen-2020-phobert}, and BARTpho \cite{DBLP:journals/corr/abs-2109-09701}.
\begin{enumerate}
    \item \textbf{ViSoBERT}  is a pre-trained language model tailored for Vietnamese social media text, trained on a 1GB dataset from platforms like Facebook, YouTube, and TikTok. It excels at interpreting NSWs such as emojis, abbreviations, and slang, using a transformer architecture and a specialized SentencePiece tokenizer, making it superior to traditional models in understanding Vietnamese online communication.
    \item \textbf{PhoBERT}, a pre-trained language model tailored for Vietnamese, is built on the RoBERTa architecture and leverages a substantial 20GB corpus from various sources including news websites, online forums, and social media to enhance NLP task performance. PhoBERT is trained using the Masked Language Modeling (MLM) method, where random tokens are masked and predicted based on surrounding context. Available in two versions, PhoBERT-base and PhoBERT-large, this study employs the base version to balance computational efficiency with model capability.
    \item \textbf{BARTpho} represents a significant advancement in Vietnamese NLP as the first pre-trained seq2seq language model tailored specifically for Vietnamese. Based on the BART architecture, it employs a two-step training process involving the introduction of random noise to input text and its reconstruction by optimizing the cross-entropy loss between the output of decoder and the original text. BARTpho is available in two versions: $\text{BARTpho}_\text{syllable}$ and $\text{BARTpho}_\text{word}$. For this study, $\text{BARTpho}_\text{syllable}$ is used due to its suitability for lexical normalization tasks.
\end{enumerate}

Pre-trained language models are renowned for their semantic representation capabilities and adaptability across various NLP tasks, including text classification, text summarization, and machine translation. For the task of lexical normalization, we leverage the power of MLM. By masking tokens that require normalization in the input, we fine-tune the model to predict these masked tokens in the output sequence. To achieve this, rather than tokenizing the input and output sentences conventionally, we perform token-level alignment tokenization to ensure the model can accurately map the positions of NSWs to their standardized counterparts. Additionally, we customize the architecture of the pre-trained models to meet the specific demands of the normalization task.

\subsubsection{Token-Level Alignment Tokenization}
Language models must predict the standard word corresponding to each NSW in the input sentence. Therefore, the positions of NSWs in the input sentence and their corresponding labels in the output sentence need to be aligned. 

Muller et al. (2019) \cite{muller-etal-2019-enhancing} proposed two tokenization algorithms for lexical normalization: Independent Alignment and Parallel Alignment. Both algorithms can tokenize sentence pairs and ensure alignment between the source and target sentences. However, these algorithms are only applicable when using a WordPiece tokenizer. This paper introduces a simpler tokenization algorithm that can be applied to any tokenizer. Specifically, the paper experiments with the PhoBERT tokenizer, which uses the Byte-Pair Encoding (BPE) method, as well as ViSoBERT and BARTpho, which utilize the SentencePiece tokenizer. Nonetheless, our tokenization algorithm imposes stricter requirements on input sentence pairs. Instead of processing input as text strings as in study \cite{muller-etal-2019-enhancing}, the source and target sentences must be provided as lists with equal lengths, where each element in the source list is aligned with the corresponding element in the target list. Before tokenization, we preprocess the input text to achieve the desired format.

\begin{figure}[ht]
    \centering
    \includegraphics[width=\linewidth]{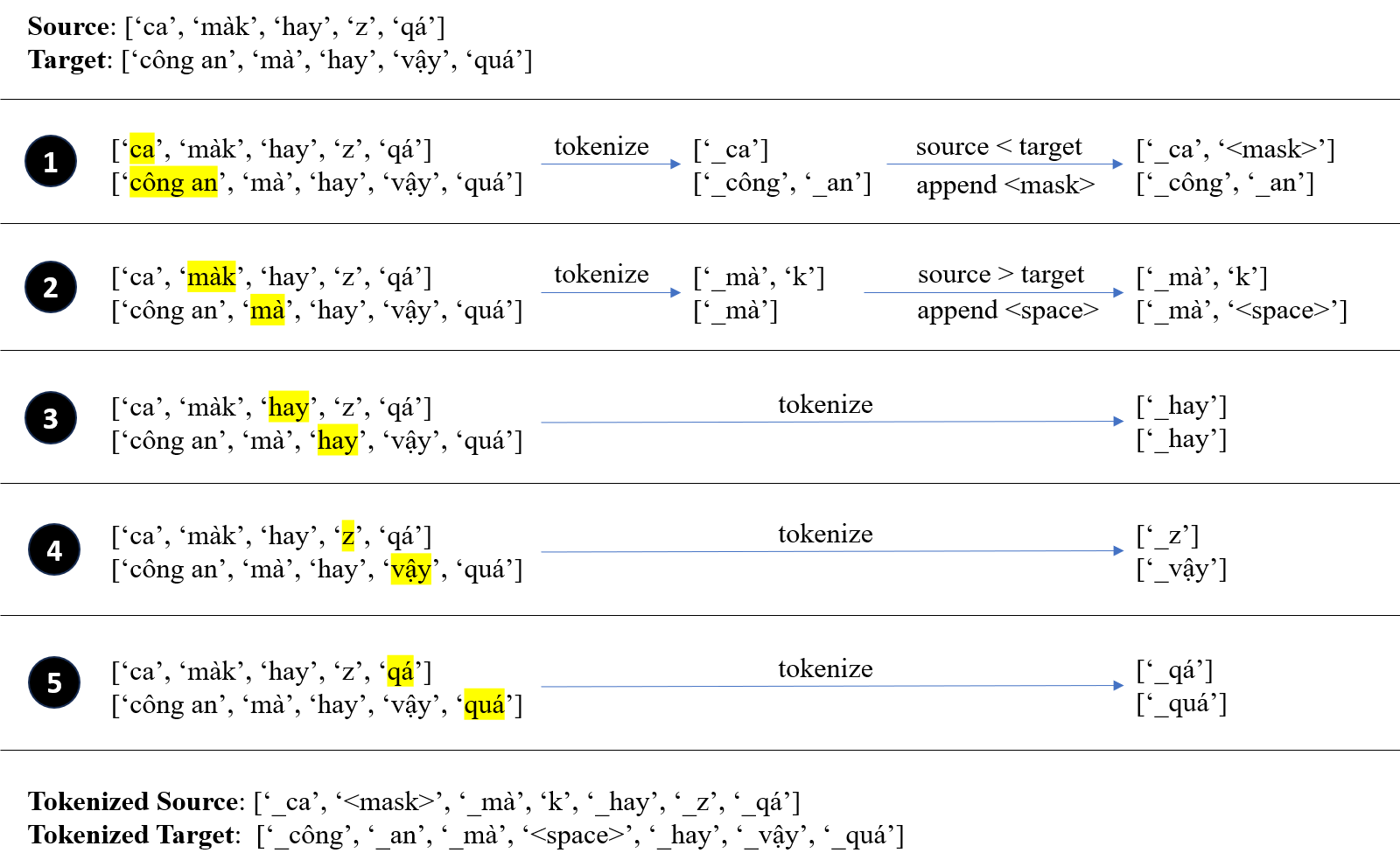}
    \caption{Token-Level Alignment Tokenization Process of the source [`ca', `màk', `hay', `z', `qá'] and target sentence [`công an', `mà', `hay', `vậy', `quá'] (\textbf{English}: The police are always like that) using ViSoBERT tokenizer.}
    \label{fig:tokenize}
\end{figure}

Our proposed tokenization algorithm is illustrated in Figure \ref{fig:tokenize}. We perform the following steps for token-level alignment tokenization:
\begin{itemize}
    \item Use a tokenizer to simultaneously tokenize both sentences.
    \item If the number of tokens in the input sentence is fewer than the number of tokens in the output sentence, we add <mask> tokens to the input token list so that the number of tokens matches on both sides.
    \item If the number of tokens in the input sentence is greater than the number of tokens in the output sentence, we add <space> tokens to the output token list so that the number of tokens matches on both sides.
\end{itemize}

\begin{table}[]
\centering
\begin{tabular}{cccccc}
\toprule
\multicolumn{2}{c}{\textbf{ViSoBERT}} & \multicolumn{2}{c}{\textbf{PhoBERT}} & \multicolumn{2}{c}{\textbf{BARTpho}} \\
\midrule
\textbf{Source}        & \textbf{Target}       & \textbf{Source}       & \textbf{Target}       & \textbf{Source}       & \textbf{Target}       \\
\midrule
\_ca          & \_công       & ca           & công         & \_ca         & \_công       \\
<mask>        & \_an         & <mask>       & an           & <mask>       & \_an         \\
\_mà          & \_mà         & mà@@         & mà           & \_mà         & \_mà         \\
k             & <space>      & k            & <space>      & k            & <space>      \\
\_hay         & \_hay        & hay          & hay          & \_hay        & \_hay        \\
\_z           & \_vậy        & z            & vậy          & \_z          & \_vậy        \\
\_qá          & \_quá        & q@@          & quá          & \_q          & \_quá        \\
              &              & á            & <space>      & á            & <space>      \\
\bottomrule
\end{tabular}
\caption{Token-Level Alignment Tokenization of `ca màk hay z qá' normalized as `công an mà hay vậy quá' (\textbf{English}: The police are always like that) using different tokenizers.}
\label{tab:aligned_tokenizer}
\end{table}

Table \ref{tab:aligned_tokenizer} illustrates the tokenization results of the source sentence [`ca', `màk', `hay', `z', `qá'] (`ca màk hay z qá') – target [`công an', `mà', `hay', `vậy', `quá'] (`công an mà hay vậy quá' - \textbf{English}: `The police are always like that') performed by the tokenizers of the ViSoBERT, PhoBERT, and BARTpho models, respectively. The insertion of additional tokens <mask> and <space> necessitates adjustments in the architecture of the pre-trained language models, as discussed in Section \ref{sect:elma}.

\subsubsection{Enhance Language Model Architecture for Lexical Normalization}
\label{sect:elma}

In the previous section, we introduced the <space> token into output sentences to facilitate the identification of corresponding tokens in input sentences that require removal. Since <space> isn't initially included in the vocabularies of ViSoBERT, PhoBERT, and BARTpho models, we expanded each vocabulary size accordingly: ViSoBERT to 15,003, PhoBERT to 64,001, and BARTpho to 43,001. Additionally, we augmented the final softmax layer with a vector initialized based on the standard distribution of the pre-trained model's existing embeddings. This adjustment ensures that the embedding of <space> aligns with the model's learned embedding space, enhancing its ability to handle instances where this new token is used.

\begin{table}[]
\centering
\begin{tabular}{|l|l|l|l|l|l|l|l|}
\hline
Source    & \_ca   & <mask> & \_mà & k     & \_hay & \_z   & \_qá  \\ \hline
Target    & \_công & \_an & \_mà & <space> & \_hay & \_vậy & \_quá \\ \hline
\#n\_mask & 1      & -1   & 0    & 0     & 0     & 0     & 0     \\ \hline
\end{tabular}
\caption{The number of next masks for each source token.}
\label{tab:n_mask}
\end{table}

Moreover, as discussed earlier, we introduced the <mask> token into input sentences to prompt the model to learn to fill in tokens and generate standardized vocabulary forms. However, incorporating <mask> introduces challenges during prediction since we don't know when to apply it. To address this, we integrated a classification module into the architecture of pre-trained model. This module uses the hidden state from the last layer of each token to predict the number of <mask> tokens needed following each token. Table \ref{tab:n_mask} illustrates an example of this prediction process. During inference, after estimating the required number of <mask> tokens, the model proceeds to predict the tokens in their standardized forms.

\subsection{Teacher Model} \label{teacher}
\subsubsection{Weak Rules}
\subsubsection*{Dictionary}
Lexical normalization using dictionaries is a widely adopted method for automating spelling error correction and standardizing text. This approach involves using dictionaries to identify NSWs and replace them with appropriate corrections.

The main advantage of this approach is its simplicity and effectiveness. By maintaining a dictionary that lists common NSWs alongside their correct counterparts, the method efficiently identifies and corrects NSWs in text. However, this method has notable limitations highlighted in previous research. It heavily relies on the quality and completeness of the dictionary used, which can lead to missed instances or inaccuracies in textual normalization. Additionally, challenges arise in handling complex language contexts where words or phrases may have multiple meanings depending on the context.

To address these challenges, our research focuses on restricting dictionary entries to 1-1 mapping words. This strategic approach aims to improve labeling accuracy and reduce confusion caused by ambiguous semantics.

\subsubsection*{Regular Expression}

Regular Expressions (RegEx) are a powerful tool in NLP, especially useful for identifying specific patterns where words deviate from standard rules.

While RegEx may not cover every word and semantic nuance as comprehensively as dictionaries, it excels in handling irregular word variations like repeated characters at the end, which dictionaries struggle with. RegEx uses wildcard characters and repetition rules to identify these patterns efficiently. For example, ``vui+i+'' detects ``vui'' with any number of ``i'' repetitions, and ``đẹpp+'' identifies ``đẹp'' with multiple ``p''s.

However, a significant challenge with RegEx is the manual construction of regular expressions. This process can be time-consuming, labor-intensive, and may not capture all nuances of natural language effectively.

\subsubsection{Rule Attention Network (RAN)}

The Rule Attention Network (RAN) enhances label assignment tasks by integrating multiple sources of weak supervision, such as regular expressions and dictionaries, each with adaptable reliability weights. This approach aims to refine the traditional methods of managing unlabeled data. Specifically, RAN aids in making more precise labeling decisions when only a limited number of weak rules are available.

Heuristic rules, like regular expressions and dictionaries, are helpful for text standardization. However, these rules frequently fail to capture a wide range of cases, leaving a significant portion of data unlabeled in conventional weak supervision approaches. To mitigate this issue, pseudo-labels predicted by a base Student model are utilized. By applying the Student model to the unlabeled dataset $D_U$, predictions $p_{\theta}(y\rvert x)$ are generated, serving as an additional source of weak supervision. This strategy increases the coverage of the weak supervision sources, thereby enhancing the overall labeling process.

For each data sample $x_i$, a set of heuristic rules $R_i$ is applied. RAN aims to integrate the weak labels derived from these heuristic rules with pseudo-labels produced by the Student model to generate a soft label $q_i$ for each data sample. To achieve this, RAN employs trainable weights to estimate and adjust the relative reliability of the different heuristic rules, thereby optimizing their adaptability.

The pseudo-label $q_i$ is computed as follows:

\begin{align}\label{eq:pseudo_label}
    q_i = \frac{1}{Z_i} \left(\sum\limits_{j: r^j \in R_i} a_i^j q_i^j + a_i^S p_{\theta}(y\rvert x) + a_i^u u\right)
\end{align}

where $a_i^j$ and $a_i^S$ represent the confidence weights for the labels $q_i^j$ predicted by rule $r^j$ and the pseudo-label $p_{\theta}(y\rvert x)$ predicted by the Student model for data sample $x_i$. The uniform distribution $u$ assigns equal probabilities to all $K$ classes, represented as $u = \left[\frac{1}{K},...,\frac{1}{K}\right]$. The weight $a_i^u$ for the uniform distribution is calculated as $a_i^u = \lvert R_i \rvert + 1 - \sum_{j: r^j \in R_i} a_i^j - a_i^S$. $Z_i$ is a normalization factor ensuring that $q_i$ is a valid probability distribution. The uniform distribution $u$ acts as a smoothing factor to prevent overfitting, especially when only a few weak rules apply to a sample.

In Equation \ref{eq:pseudo_label}, a rule $r^j$ with a higher confidence weight $a^j$ will contribute more to $q_i$. If $a^j = 1$ for all $r^j \in R_i \cup p_{\theta}$, RAN simplifies to a majority voting scheme. Conversely, if $a^j = 0$ for all $r^j \in R_i \cup p_{\theta}$, RAN predicts $q_i = u$.

Typically, heuristic rules are predefined and often operate based on simple patterns. For example, a dictionary may dictate that the abbreviation `ca' should be translated to the standard form $\text{`công an'}_{\text{police}}$, but in some cases, $\text{`ca'}_{\text{cup}}$ should remain unchanged. Thus, these rules often have limitations and cannot capture the complex nature of the data. To address this issue, RAN leverages embeddings of data samples. Embeddings are dense vector representations of input sentences, capable of capturing their semantics. These embeddings are usually generated by neural network models, in this case, the Student model.

By utilizing embeddings, RAN can understand the context in which rules should be applied to the input sentence. Moreover, rules are also represented by embeddings, allowing the model to learn semantic similarities between rules and specific sentences. RAN employs an attention mechanism to compute how much attention should be placed on each rule for each input sentence. Specifically, RAN utilizes the hidden state representation of a data sample $x_i$ as $h_i \in \mathbb{R}^{d_o}$, then applies a multi-layer perceptron (MLP) network to map this embedding to the same space as rule embeddings ($e_j = g(r^j) \in \mathbb{R}^d$). Subsequently, the attention weight $a_i^j$ is computed using a sigmoid activation function to determine the relevance of each rule to $x_i$, as formulated in Equation \ref{eq:attention_w}. This mechanism allows RAN to dynamically adjust the attention placed on each rule based on the context of the input sentence $x_i$, enhancing its ability to effectively utilize weak supervision sources for label assignment.

\begin{align}\label{eq:attention_w}
    a_i^j = \sigma\left(f(h_i)^T \cdot e_j\right) \in [0, 1]
\end{align}

In Equation \ref{eq:attention_w}, $f$ is an MLP that maps $h_i$ into the space of $e_j$, and $\sigma$ is the sigmoid activation function, ensuring that the attention weights lie within the interval [0,1]. Rule embeddings enable the Rule Attention Network (RAN) to exploit similarities between different rules and leverage their semantics to model consensus. The attention weight of the Student model, $a_i^S$, is calculated similarly to the weights of heuristic rules.

Training the RAN involves optimizing the attention weights so that the combined labels $q_i$ achieve the highest possible accuracy. The model learns to assign higher weights to more reliable rules and lower weights to less reliable ones, all based on the specific context and semantics of each sentence.

In summary, the labels $q_i$ are considered fixed, while their attention weights are estimated. The connection between rules and data samples through embeddings $e_j$ and $h_i$ allows RAN to model consensus via attention weights $a_i^j$. The trainable parameters of RAN (i.e., $f$ and $g$) are shared across all rules and data samples, forming a robust and flexible weakly supervised learning approach.

\begin{figure}[ht]
    \centering
    \includegraphics[width=\linewidth]{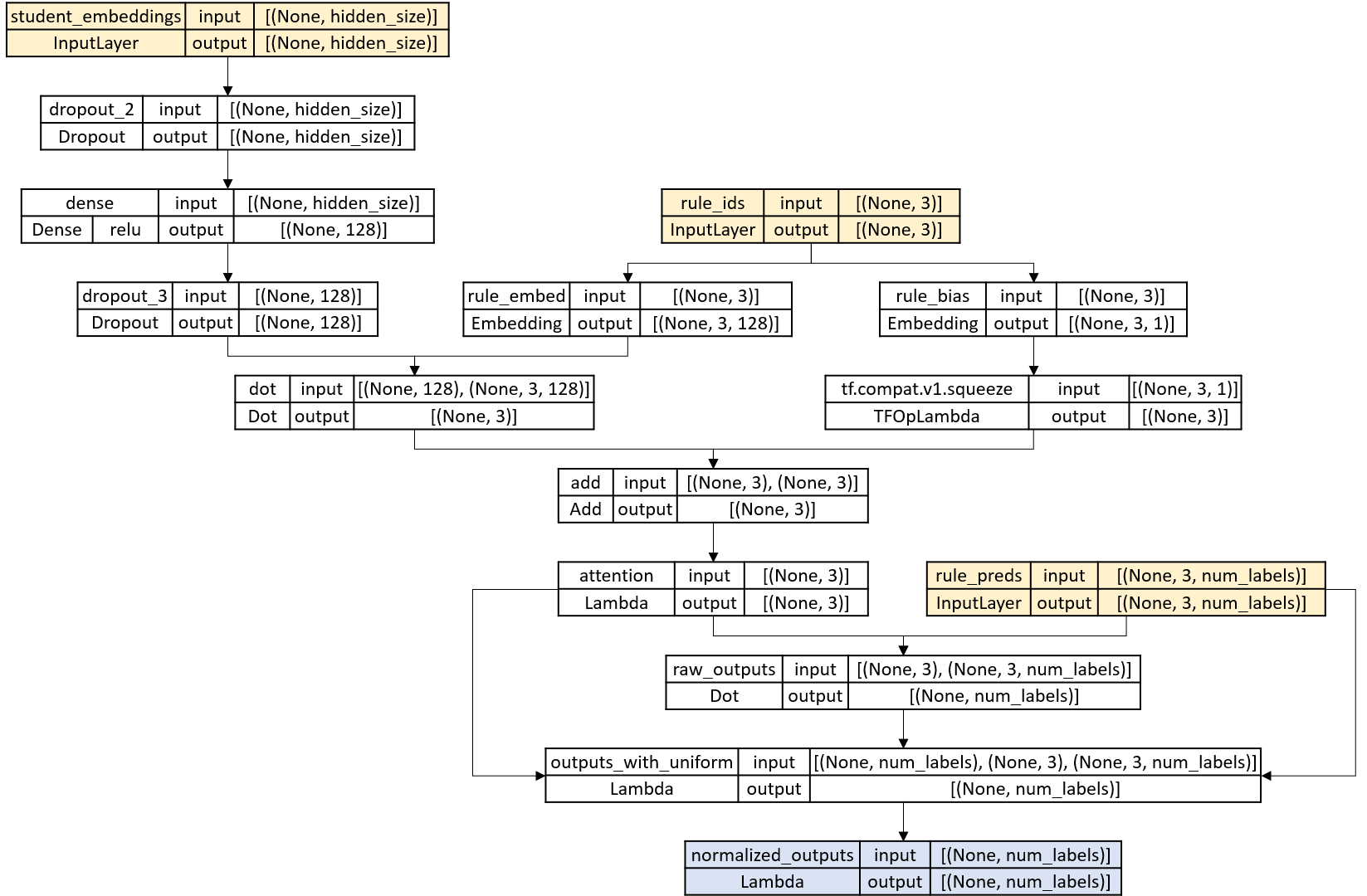}
    \caption{The Architecture of Rule Attention Network.}
    \label{fig:ran}
\end{figure}

Figure \ref{fig:ran} presents the architecture of RAN, highlighting its three primary inputs:
\begin{itemize}
    \item \textbf{student\_embeddings}: These are vector representations of the input text generated by the Student model. The dimensionality of the vectors varies depending on the specific model: PhoBERT and ViSoBERT produce vectors with 768 dimensions, while BARTpho yields vectors with 1024 dimensions.
    \item \textbf{rule\_ids}: This tensor encodes the indices of the rules applied to each token. There are three rules: index 0 corresponds to a regular expression, index 1 to a dictionary, and index 2 to the Student model.
    \item \textbf{rule\_preds}: This tensor represents the normalized outputs from the three weak rules. The number of labels varies across models: 15,003 for ViSoBERT, 64,001 for PhoBERT, and 43,001 for BARTpho.
\end{itemize}
  
\subsection{Framework Training Procedure} \label{train_framework}

Before training the labeling framework, it is essential to prepare a dataset containing weak labels generated by regular expressions and dictionaries. These labels remain fixed throughout the training process, enabling the framework to learn and adjust the weight of each rule based on individual data samples. Specifically, two columns, `regex\_rule' and `dict\_rule', are added to the datasets $D_L$ and $D_U$, containing the predictions from regular expressions and dictionaries, respectively. Table \ref{tab:weak_rules} illustrates examples of these columns.

\begin{table}[ht]
\centering
\begin{tabular}{|l|l|}
\hline
\textbf{regex\_rule}                                                                                                        & \textbf{dict\_rule}                                                                                                                \\ \hline
{[}`thích', `anh', `cá mập', `không'{]}                                                                                     & {[}`thích', `anh', `cá mập', `hkk'{]}                                                                                              \\ \hline
{[}`cứ', `ngây thơ', `thế', `thoai', `:))'{]}                                                                               & {[}`cứ', `ngây thơ', `thế', `thôi', `:))'{]}                                                                                       \\ \hline
\begin{tabular}[c]{@{}l@{}}{[}`đọc', `jd', `mà', `nó', `dễ', `quá', \\ `đâm ra', `sợ', `cty', `lừa', `:)))'{]}\end{tabular} & \begin{tabular}[c]{@{}l@{}}{[}`đọc', `jd', `mà', `nó', `dễ', `quá', \\ `đâm ra', `sợ', `công ty', `lừa', `:)))'{]}\end{tabular} \\ \hline
\end{tabular}
\caption{Illustration of two additional columns in the dataset $D_L$.}
\label{tab:weak_rules}
\end{table}

\begin{figure}[ht]
    \centering
    \includegraphics[width=\linewidth]{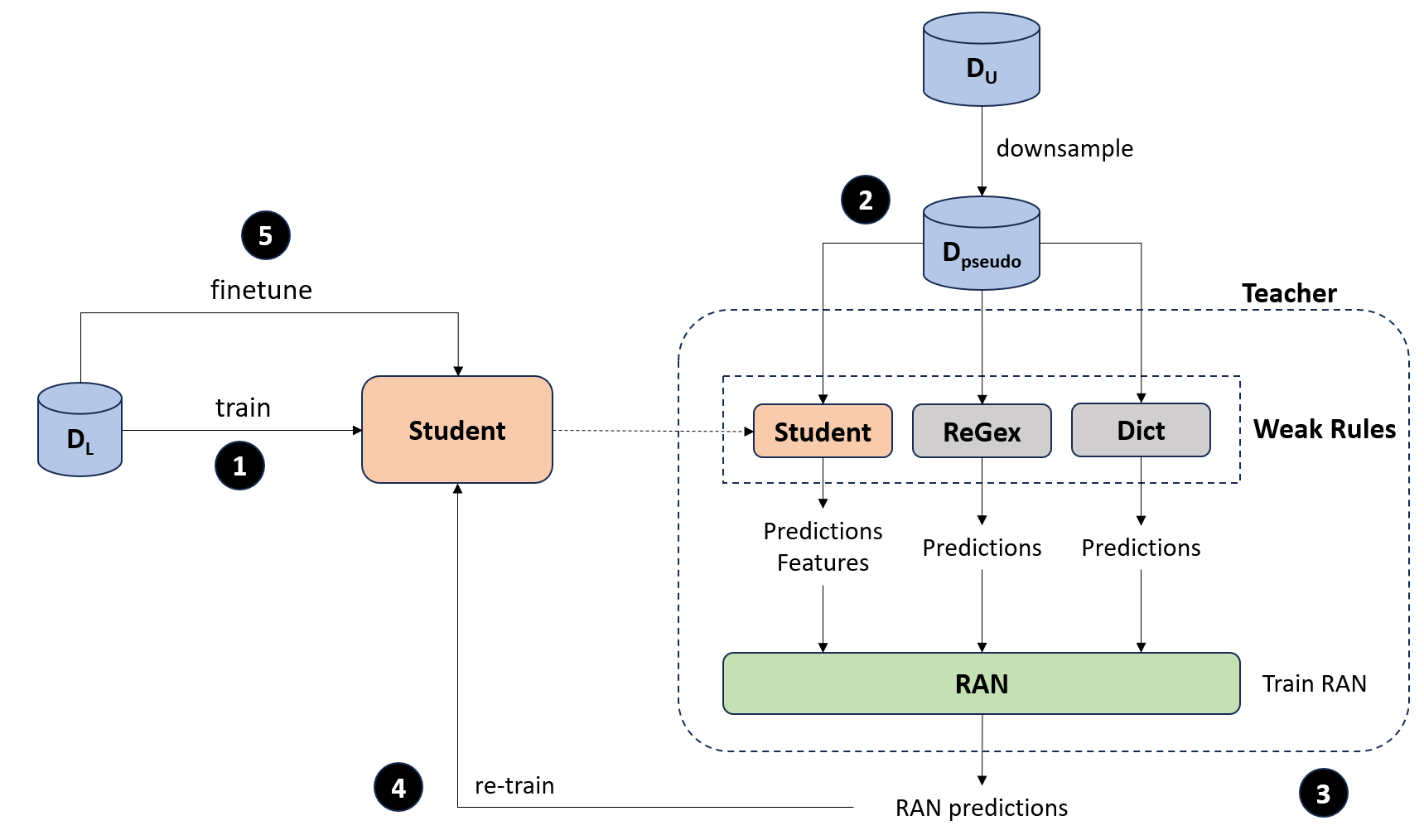}
    \caption{Weak Supervision Framework Traning Procedure.}
    \label{fig:framework_training}
\end{figure}

The training process for the complete weak supervision framework, depicted in Figure \ref{fig:framework_training}, involves five key steps: (1) training the Student model, (2) predicting pseudo-labels for $D_U$, (3) training the Rule Attention Network (RAN), (4) retraining the Student model with the $D_U$ dataset, and (5) fine-tuning the Student model with the labeled dataset $D_L$. Steps 2 through 5 are repeated iteratively for a predetermined number of iterations.

\textbf{Step 1: Traning Student model.} The initial phase involves training the Student model on the labeled dataset $D_L$ using supervised learning techniques. Upon completion, the model weights are saved, and the model's performance is evaluated on the development and test sets derived from $D_L$.

\textbf{Step 2: Predicting pseudo-labels for $D_U$ dataset.} In this step, pseudo-labels are predicted for the unlabeled dataset $D_U$. Given the large size of $D_U$ and limited training resources, a random subset $D_{pseudo}$ is first sampled from $D_U$ for prediction. The Student model then generates pseudo-labels for $D_{pseudo}$ while also extracting input data representations (embeddings), which will be used to train the RAN in the next step. The pseudo-labels predicted by regular expressions and the dictionary are directly sourced from the pre-constructed dataset.

\textbf{Step 3: RAN Training.} The RAN is trained using a combination of unsupervised and supervised learning methods to fully exploit the datasets $D_L$ and $D_U$. During the unsupervised learning phase, the Minimum Entropy loss function (Equation \ref{eq:unsup_loss}) is employed, which encourages the RAN to predict weights that maximize agreement among the weak rules. This loss function penalizes dissimilar predictions from the rules and rewards similar ones, enabling the model to assign accurate labels based solely on the input text properties and the reference labels from the weak rules, without needing true labels.

\begin{align}\label{eq:unsup_loss}
    \mathcal{L}_{unsup} = - \sum\limits_{x_i \in D_U} q_i \log q_i
\end{align}

In the supervised learning phase, the model is fine-tuned on the labeled dataset $D_L$ using the Cross-Entropy loss function, as formulated in Equation \ref{eq:sup_loss}. Here, $q_i$ represents the pseudo-label predicted by the Teacher model, and $y_i$ is the true label. Fine-tuning on the labeled dataset enhances the model's predictive accuracy. After training, the weights of RAN are saved, and its prediction results are evaluated on the development and test sets. The trained RAN is then used to predict soft labels for the $D_{pseudo}$ dataset.

\begin{align}\label{eq:sup_loss}
    \mathcal{L}_{sup} = - \sum\limits_{(x_i, y_i) \in D_L} y_i \log q_i
\end{align}

\textbf{Step 4: Re-trainning Student model with $D_{pseudo}$ dataset.} The Student model is retrained on the $D_{pseudo}$ dataset using supervised learning, with the soft labels predicted by the Teacher model serving as the labels. Unlike hard labels, which indicate a specific class for a data sample, soft labels represent the probability distribution of the prediction across all classes. Each element in a soft label vector indicates the probability of the data sample belonging to a particular class. Using predicted probabilities as labels incorporates information about prediction uncertainty, leading to a model with more stable accuracy and better generalization.

The loss function is calculated as follows:

\begin{align}\label{eq:retrain_loss}
    \mathcal{L}_{pseudo} = - \sum\limits_{i=1}^{N} \sum\limits_{j=1}^{C} q_{ij} \log p_{ij}
\end{align}

where $N$ is the number of data samples, $C$ is the number of labels, $q_{ij}$ represents the soft label (probability) of the i-th data sample belonging to the j-th class as predicted by RAN. Similarly, $p_{ij}$ denotes the probability predicted by the Student model.

\textbf{Step 5: Fine-tuning Student model on $D_L$ dataset.} Finally, the Student model is fine-tuned on the labeled dataset $D_L$. While pseudo-labels facilitate leveraging unlabeled data for model training, they may introduce noise that can impact the model's predictive capability. Fine-tuning on the labeled dataset allows the model to correct inaccurate predictions and reduce potential noise introduced by pseudo-labels, thereby improving overall performance.

\section{Experiments and Results} \label{section-5}
\subsection{Experimental Settings}

The experiments in this paper were conducted using Colab Pro with GPU L4. Specifically, experiments involving the PhoBERT and BARTpho models were executed using Colab Pro with GPU A100.

\subsubsection{Experiment 1: Evaluating the Labeling Capability of the Proposed Framework}

Our main experiment is designed to assess the labeling capability of the weakly supervised labeling framework. Labeling capability encompasses the ability to normalize words requiring normalization, maintain vocabulary that does not require normalization, and the accuracy of the entire sentence post-normalization. Specific metrics will be presented in Section \ref{subsec:metrics}.

We will sequentially run the framework with three different Student models: ViSoBERT, PhoBERT, and BARTpho. Concurrently, we will compare the performance of framework against two baselines:

\begin{enumerate}
    \item \textbf{Using Only the Student Model}: This involves training the Student model on the training set without any additional configurations.
    \item \textbf{Traditional Self-training}: This involves training the Student model using supervised learning on the training set, then using the trained model to generate pseudo-labels for the $D_U$ set. The pseudo-labels are directly added to the training set, and the Student model continues training iteratively.
\end{enumerate}

\begin{table}[]
\centering
\begin{tabular}{ccc}
\toprule
\textbf{Model}            & \textbf{Layer}     & \textbf{Learning rate} \\
\midrule
\multirow{5}{*}{ViSoBERT} & embeddings         & 5e-5                   \\
                          & encoder            & 2e-5                   \\
                          & pooler             & 1e-5                   \\
                          & classifier         & 1e-5                   \\
                          & mask\_n\_predictor & 1e-5                   \\
\midrule
\multirow{5}{*}{PhoBERT}  & embeddings         & 5e-5                   \\
                          & encoder            & 2e-5                   \\
                          & pooler             & 1e-5                   \\
                          & classifier         & 1e-5                   \\
                          & mask\_n\_predictor & 1e-5                   \\
\midrule
\multirow{5}{*}{BARTpho}  & shared             & 5e-5                   \\
                          & encoder            & 2e-5                   \\
                          & decoder            & 1e-5                   \\
                          & classifier         & 1e-5                   \\
                          & mask\_n\_predictor & 1e-5                   \\
\bottomrule
\end{tabular}
\caption{Layer-specific Learning Rates.}
\label{tab:lr}
\end{table}

In all experiments involving the training of the Student model, the Adam optimizer, known for its robustness and widespread use, is employed. The learning rates are specifically tailored for different layers of each model, with detailed values provided in Table \ref{tab:lr}.

Within the proposed framework, the RAN model is also trained using the Adam optimizer. A learning rate scheduler is utilized, starting with a gradual increase to a maximum learning rate of $1 \times 10^{-2}$ during the warmup phase, and then following an exponential decay to a minimum learning rate of $1 \times 10^{-5}$ during the decay phase.

Additional hyperparameters used in our experiments are presented in Table \ref{tab:hyperparam}.

\begin{table}[]
\centering
\begin{tabular}{lccc}
\toprule
\textbf{Hyperparameters}       & \textbf{ViSoBERT} & \textbf{PhoBERT} & \textbf{BARTpho} \\
\midrule
\multicolumn{4}{l}{\textbf{Student}}                                                     \\
\midrule
training epochs                & 10                & 10               & 10               \\
batch size                     & 16                & 16               & 16               \\
\midrule
\multicolumn{4}{l}{\textbf{Self-training}}                                               \\
\midrule
student training epochs        & 5                 & 5                & 5                \\
pseudo training epochs         & 5                 & 5                & 5                \\
training batch size            & 16                & 16               & 16               \\
unsupervised batch size        & 128               & 16               & 16               \\
self-training iterations       & 10                & 10               & 10               \\
$n_{\text{downsample}}$        & 8096              & 1000             & 1000             \\
instance vector dimensionality & 768               & 768              & 1024             \\
\midrule
\multicolumn{4}{l}{\textbf{Weakly Supervised (Ours)}}                                \\
\midrule
student training epochs        & 5                 & 5                & 5                \\
pseudo training epochs         & 5                 & 5                & 5                \\
fine-tuning epochs             & 5                 & 5                & 5                \\
self-training iterations       & 10                & 10               & 10               \\
training batch size            & 16                & 16               & 16               \\
unsupervised batch size        & 128               & 16               & 16               \\
$n_{\text{downsample}}$        & 8096              & 1000             & 1000             \\
rule embedding dimensionality  & 128               & 128              & 128              \\ 
\bottomrule
\end{tabular}
\caption{Hyperparameters Selection.}
\label{tab:hyperparam}
\end{table}

\subsubsection{Experiment 2: Investigating Diacritics Removal Ratios in Training Data to Enhance Sentence Normalization Capabilities}

In addition to our primary experiment, we also augment the training dataset to improve the framework’s ability to normalize sentences lacking diacritics (e.g., `dang o tieu vuong quoc ma an noi kieu day ha' → `đang ở tiểu vương quốc mà ăn nói kiểu đấy hả' - \textbf{English}: `Even though you're in the emirate, you talk like that'). Specifically, we replicate the training and development sets and remove diacritics from a proportion $p$ of the characters in each sentence within the newly created replicas. The experimental values for $p$ are 1, 0.8, and 0.5. We will compare the framework’s performance across these three $p$ values to determine the most optimized ratio.

\subsection{Evaluation Metrics} \label{subsec:metrics}
\subsubsection{Metrics for Evaluating Vocabulary Normalization Accuracy}
Let $\text{\#need\_norm}$ represent the number of words that require normalization, $\text{\#pred\_need\_norm}$ represent the number of words for which the model performs normalization (i.e., it provides a prediction different from the word in the input sentence), and $\text{TP}_{\text{need\_norm}}$ represent the number of $\text{\#need\_norm}$ words that are correctly predicted. Recall, Precision, and F1-score metrics are defined as follows.

\begin{align}\label{eq:recall}
    \text{recall} = \frac{\text{TP}_{\text{need\_norm}}}{\text{\#need\_norm}}
\end{align}

\begin{align}\label{eq:precision}
    \text{precision} = \frac{\text{TP}_{\text{need\_norm}}}{\text{\#pred\_need\_norm}}
\end{align}

\begin{align}\label{eq:f1-score}
    \text{f1-score} = \frac{2 \times \text{recall} \times \text{precision}}{\text{recall} + \text{precision}}
\end{align}

\subsubsection{Metrics for Evaluating the Integrity of Words Not Requiring Normalization}
During the prediction process, words that do not require normalization should remain unchanged. However, the model may inadvertently alter these words, contradicting the objective of the problem. Thus, evaluating the integrity of words that do not require normalization is also crucial.

Let $\text{\#need\_no\_norm}$ denote the words that do not require normalization (and should remain unchanged during prediction), and $\text{TP}_{\text{need\_no\_norm}}$ denote the number of $\text{\#need\_no\_norm}$ words correctly predicted as unchanged. The metric for evaluating the integrity of words that do not require normalization is defined as follows:

\begin{align}\label{eq:integrity}
    \text{Integrity Score} = \frac{\text{TP}_{\text{need\_no\_norm}}}{\text{\#need\_no\_norm}}
\end{align}

\subsubsection{Sentence-Level Accuracy Metric}
Let $n_{\text{token}}$ denote the number of tokens in a sentence, and $TP$ denote the number of tokens in the predicted sentence that match the target sentence. The accuracy metric is defined as follows:

\begin{align}\label{eq:accuracy}
    \text{accuracy} = \frac{TP}{n_{\text{token}}}
\end{align}
\subsection{Analysis and Discussion}
\subsubsection{Experiment 1}

\begin{table}[]
\centering
\begin{tabular}{ccccc}
\toprule
\textbf{Metric}                                                           & \textbf{Base Model} & \textbf{Student} & \textbf{Self-Training} & \textbf{\begin{tabular}[c]{@{}c@{}}Weakly Supervised\\ (Ours)\end{tabular}} \\
\midrule
\multirow{3}{*}{\begin{tabular}[c]{@{}c@{}}Precision\\ (\%)\end{tabular}} & ViSoBERT            & 74.25            & 72.55                  & \textbf{75.91}                                                              \\
                                                                          & PhoBERT             & 71.39            & 73.21                  & \textbf{74.24}                                                              \\
                                                                          & BARTpho             & 87.52            & 88.15                  & \textbf{88.30}                                                              \\
\midrule
\multirow{3}{*}{\begin{tabular}[c]{@{}c@{}}Recall\\ (\%)\end{tabular}}    & ViSoBERT            & 75.52            & 76.23                  & \textbf{76.85}                                                              \\
                                                                          & PhoBERT             & 71.60            & 75.15                  & \textbf{75.98}                                                              \\
                                                                          & BARTpho             & 78.94            & 82.91                  & \textbf{83.32}                                                              \\
\midrule
\multirow{3}{*}{\begin{tabular}[c]{@{}c@{}}F1-score\\ (\%)\end{tabular}}  & ViSoBERT            & 73.97            & 73.56                  & \textbf{75.79}                                                              \\
                                                                          & PhoBERT             & 70.89            & 73.71                  & \textbf{75.44}                                                              \\
                                                                          & BARTpho             & 82.72            & 84.79                  & \textbf{84.94}                                                              \\
\midrule
\multirow{3}{*}{\begin{tabular}[c]{@{}c@{}}Accuracy\\ (\%)\end{tabular}}  & ViSoBERT            & 95.17            & 94.84                  & \textbf{95.42}                                                              \\
                                                                          & PhoBERT             & 92.51            & 93.08                  & \textbf{94.40}                                                              \\
                                                                          & BARTpho             & 95.48            & 95.27                  & \textbf{96.06}                    \\
\bottomrule
\end{tabular}
\caption{Comparative Results of Text Normalization Using Weakly-Supervised Framework and Baselines.}
\label{tab:exp1-1}
\end{table}

Table \ref{tab:exp1-1} presents a performance comparison of the ViSoBERT, PhoBERT, and BARTpho models when applying various methods, including using only a base model (Student), self-training, and the weakly-supervised training for lexical normalization. The results are evaluated based on Precision, Recall, F1-score, and Accuracy metrics.

The results presented in the table highlight the robustness of the proposed weakly supervised framework across various metrics and models. Notably, the framework consistently outperforms other methods in Precision, Recall, F1-score, and Accuracy. For instance, the weakly supervised approach achieves the highest Precision for ViSoBERT at 75.91\% and PhoBERT at 74.24\%, surpassing both the Student and Self-Training approaches. Additionally, it delivers an exceptional F1-score for BARTpho at 84.94\%, and the highest Accuracy across all models, particularly 96.06\% for BARTpho. These improvements demonstrate the effectiveness of the weakly supervised method in enhancing model performance through better utilization of unlabeled data.

BARTpho leads in Precision, Recall, and F1-score, achieving 88.52\%, 78.94\%, and 82.72\% respectively, which is significantly higher than ViSoBERT and PhoBERT. Thus, BARTpho has proven its advantage in lexical normalization, affirming its efficacy in processing and normalizing natural language in real-world applications. ViSoBERT is designed for social media language, giving it a slight edge in text normalization results compared to PhoBERT. However, its vocabulary size and pre-trained dataset are much smaller than those of the other two models, making it difficult to surpass BARTpho in normalization accuracy.

\begin{table}[]
\centering
\begin{tabular}{cccc}
\toprule
\textbf{Base Model} & \textbf{Student} & \textbf{Self-Training} & \textbf{Weakly Supervised (Ours)} \\
\midrule
ViSoBERT       & 98.25            & 97.79                  & 98.26                       \\
PhoBERT        & 97.08            & 96.99                  & 97.38                       \\
BARTpho        & 99.22            & 98.97                  & 99.04                       \\
\bottomrule
\end{tabular}
\caption{Comparison of Integrity (\%) of Canonical Words.}
\label{tab:exp1-2}
\end{table}

Table \ref{tab:exp1-2} compares the integrity of words that do not require normalization, showing a clear advantage for the BARTpho model, with integrity rates of 99.22\%, 98.97\%, and 99.04\% for the Student, Self-training, and Weakly-supervised Framework methods respectively. This demonstrates that BARTpho maintains the originality of words that do not need normalization with the highest accuracy. Meanwhile, ViSoBERT and PhoBERT also show relatively high performance, with integrity rates of 98.25\% and 97.08\% for ViSoBERT, and 97.79\% and 96.99\% for PhoBERT.

Among the three baselines, self-training exhibits the lowest capability to maintain the integrity of canonical  words. The weakly-supervised framework’s ability in this aspect is also slightly reduced compared to using only a single Student model for lexical normalization. This may be due to the accuracy of the self-training and weakly-supervised training models being influenced by pseudo-labels. The weakly-supervised framework is less affected as it is fine-tuned on the correct labels.
\subsubsection{Experiment 2}

\begin{table}[]
\centering
\begin{tabular}{ccccc}
\toprule
\textbf{Base Model}            & \textbf{$p$} & \textbf{F1-score (\%)} & \textbf{Integrity Score (\%)} & \textbf{Accuracy (\%)} \\
\midrule
\multirow{4}{*}{ViSoBERT} & 0        & 75.79                  & 98.26                         & 95.42                  \\
                          & 0.5       & 71.82                  & 96.43                         & 93.95                  \\
                          & 0.8       & 70.86                  & 96.62                         & 93.99                  \\
                          & 1      & 72.19                  & 96.91                         & 94.42                  \\
\midrule
\multirow{4}{*}{BARTpho}  & 0        & 84.94                  & 99.04                         & 96.06                  \\
                          & 0.5       & 84.27                  & 98.2                          & 95.85                  \\
                          & 0.8       & 85.4                   & 98.3                          & 96.04                  \\
                          & 1      & 85.64                  & 98.61                         & 96.16                  \\
\bottomrule
\end{tabular}
\caption{Comparative Results of Text Normalization Using ViSoBERT and BARTpho with Different Diacritics Removal Ratios ($p$).}
\label{tab:diacritic}
\end{table}

Table \ref{tab:diacritic} presents the text normalization results of ViSoBERT and BARTpho when applying different diacritics removal rates ($p$), aimed at evaluating the models' performance in handling text with and without diacritics.

For ViSoBERT, as the diacritics dropping rate increases, the model's performance noticeably decreases. Specifically, when $p$ = 50\%, the F1-score decreases to 71.82\%, integrity score drops to 96.43\%, and Accuracy decreases to 93.95\%. Similarly, at p = 80\%, all three metrics show a decrease. When all tokens are dropped ($p$ = 100\%), the F1-score slightly improves to 72.19\%, integrity score reaches 96.91\%, and Accuracy is 94.42\%. These numbers indicate that ViSoBERT struggles more with lexical normalization as the diacritics dropping rate increases, but shows slight recovery when all diacritics are dropped.

In contrast, BARTpho demonstrates stability and superior performance under the same conditions. As the diacritics dropping rate increases, BARTpho maintains high performance levels. Specifically, at $p$ = 50\%, the F1-score is 84.27\%; at $p$ = 80\%, the F1-score increases to 85.40\%, and Accuracy is 96.04\%. When all tokens are dropped ($p$ = 100\%), the F1-score continues to improve to 85.64\%, and Accuracy reaches 96.16\%. This indicates that BARTpho effectively handles text with and without diacritics, maintaining high performance even as the diacritics dropping rate increases. However, it is noted that as $p$ increases, while BARTpho becomes stronger in lexical normalization, its ability to preserve canonical words decreases, reflected in the integrity score being lower at $p$ = 50\%, 80\%, and 100\% compared to $p$ = 0\%.

Overall, BARTpho demonstrates superior and stable performance compared to ViSoBERT in text normalization across different token dropping rates, highlighting its flexibility and robustness in handling diverse language processing conditions.

\subsection{Errors Analysis}
\subsubsection{Errors in Predictions of $D_L$ dataset}

\begin{table}[]
\centering
\begin{tabular}{ll}
\toprule
\textbf{Sentence ID}                 & 1                                                               \\
\textbf{Source}                      & đh gia đình chưa cho đi làm :))                                 \\
\textbf{Target}                      & sinh viên đại học gia đình chưa cho đi làm :))                  \\
\textbf{English}                     & university student whose family has not allowed to work yet :)) \\
\multirow{2}{*}{\textbf{Prediction}} & sinh học đại học gia đình chưa cho đi làm :)) (\textbf{visobert})        \\
                                     & sinh viên đại học gia đình chưa cho đi làm (\textbf{bartpho})            \\
\midrule
\textbf{Sentence ID}                 & 2                                                               \\
\textbf{Source}                      & các founder có nhận fb ko                                       \\
\textbf{Target}                      & các founder có nhận fb không                                    \\
\textbf{English}                     & do the founders accept facebook                                 \\
\multirow{2}{*}{\textbf{Prediction}} & các f thậter có nhận fb không (\textbf{visobert})                        \\
                                     & các founder có nhận fb không (\textbf{bartpho})                         \\
\midrule
\textbf{Sentence ID}                 & 3                                                               \\
\textbf{Source}                      & nhìn noá lại nghĩ đến con ngỗng nhà toy                         \\
\textbf{Target}                      & nhìn nó lại nghĩ đến con ngỗng nhà tôi                          \\
\textbf{English}                     & looking at it, i think of my goose                              \\
\multirow{2}{*}{\textbf{Prediction}} & nhìn no lại nghĩ đến con nhà tôi (\textbf{visobert})                     \\
                                     & nhìn nó lại nghĩ đến con ngỗng nhà tôi (\textbf{bartpho})               \\
\bottomrule
\end{tabular}
\caption{Several Prediction Instances from $D_L$: Overall Lexical Normalization Capability.}
\label{tab:error_lexnorm}
\end{table}

Table \ref{tab:error_lexnorm} illustrates several results of lexical normalization using the weakly supervised framework with two base models: ViSoBERT and BARTpho. We will analyze this table to highlight some common errors in the lexical normalization process:

\begin{itemize}
    \item \textbf{Analysis of Example 1}: The words requiring normalization in this example are ‘sv’ ($\text{`sinh viên'}_\text{student}$) and ‘đh’ ($\text{`đại học'}_\text{university}$). It can be observed that the ViSoBERT model consistently provides incorrect normalization ($\text{`sinh học'}_\text{biology}$), while the BARTpho model correctly predicts the normalized forms. However, a drawback of the BARTpho model is its inability to encode emoticons, resulting in the omission of the emoticon at the end of the sentence in the output of BARTpho.
    \item \textbf{Analysis of Example 2}: When encountering words in other languages, the ViSoBERT model tends to incorrectly modify these words, whereas the BARTpho model demonstrates greater stability, though occasional errors still occur.
    \item \textbf{Analysis of Example 3}: In this example, the words requiring normalization are ‘nóa’ ($\text{`nó'}_\text{it}$) and ‘toy’ ($\text{`tôi'}_\text{I}$). The ViSoBERT model fails to normalize the word `nóa' but successfully normalizes `toy'. In contrast, the BARTpho model handles both cases effectively. An important detail in this example is that the word $\text{`ngỗng'}_\text{goose}$ is a proper term and should not be altered, and BARTpho successfully retains it unchanged. However, ViSoBERT erroneously removes the word `ngỗng' from the sentence. This example clearly reflects a limitation of ViSoBERT, namely that its vocabulary size is too small and does not include the word `ngỗng'.
\end{itemize}

Briefly, in lexical normalization, common errors include incorrect normalization of abbreviations, loss of contextual elements such as emoticons, and over-correction of foreign words. Additionally, limited vocabulary coverage can lead to improper handling or removal of rare terms, underscoring the need for richer vocabularies and improved training methodologies.

\begin{table}[]
\centering
\begin{tabular}{ll}
\hline
\toprule
\textbf{Sentence ID}                                                                      & 1                                                                       \\
\textbf{Source}                                                                           & dang o tieu vuong quoc ma an noi kieu day ha                            \\
\textbf{Target}                                                                           & đang ở tiểu vương quốc mà ăn nói kiểu đấy hả                            \\
\textbf{English}                                                                          & even though you're in the emirate, you talk like that                   \\
\midrule
\multirow{4}{*}{\textbf{\begin{tabular}[c]{@{}l@{}}Prediction\\ (ViSoBERT)\end{tabular}}} & đang vậy trường mà ăn nói dạy ha (p = 0)                                \\
                                                                                          & đang ở tiêu vậy quốc mà ăn nói kiểu dạy hả (p = 0.5)                    \\
                                                                                          & đang ở tiêu quốc mà ăn nói kiểu đây hả (p = 0.8)                        \\
                                                                                          & đang ở tiểu vương quốc má ăn nội kiểu đây hả (p = 1)                    \\
\midrule
\multirow{4}{*}{\textbf{\begin{tabular}[c]{@{}l@{}}Prediction\\ (BARTpho)\end{tabular}}}  & đang o tui hưởng trước mà an nói ngu dạy ha (p = 0)                     \\
                                                                                          & đang  tiểu quốc mà ăn nổi kiểu vậy ha (p = 0.5)                         \\
                                                                                          & đang ở tiểu quốc vượt quốc quốc mà ăn nói kiểu đây hả (p = 0.8)         \\
                                                                                          & đang ở tiểu trưởng quốc mà ăn nói kiểu đây hả (p = 1)                   \\
\hline
\toprule
\textbf{Sentence ID}                                                                      & 2                                                                       \\
\textbf{Source}                                                                           & vn nói ``hết giai tao đợi mày ơ sân bay"                                 \\
\textbf{Target}                                                                           & việt nam nói ``hết giải tao đợi mày ở sân bay"                           \\
\textbf{English}                                                                          & vietnam said ``after the tournament, i will wait for you at the airport" \\
\midrule
\multirow{4}{*}{\textbf{\begin{tabular}[c]{@{}l@{}}Prediction\\ (ViSoBERT)\end{tabular}}} & việt nói ``hết giai tao đợi mày ơ sân bây" (p = 0)                       \\
                                                                                          & việt nói ``hết giải tao đợi mày ơ sân bây" (p = 0.5)                     \\
                                                                                          & việt nói ``hết giải tao đợi mày ơ sân bây" (p = 0.8)                     \\
                                                                                          & vn nói ``hết giải tao đợi mày ơ sân bay" (p = 1)                         \\
\midrule
\multirow{4}{*}{\textbf{\begin{tabular}[c]{@{}l@{}}Prediction\\ (BARTpho)\end{tabular}}}  & việt nam nói ``hết trai tao đợi mày ơ sân bay" (p = 0)                   \\
                                                                                          & việt nam nói ``hết giải tao đợi mày ơ sân bay" (p = 0.5)                 \\
                                                                                          & v nam nói ``hết trai tao đợi mày ởơ sân bay" (p = 0.8)                   \\
                                                                                          & việt nam nói ``hết trai tao đợi mày ơ sân bay" (p = 1)                  \\
\bottomrule
\hline
\end{tabular}
\caption{Several Prediction Instances from $D_L$: Lexical Normalization with Diacritics Removal.}
\label{tab:error_diacritics}
\end{table}

Table \ref{tab:error_diacritics} illustrates several prediction examples from the weakly supervised framework using two base models, ViSoBERT and BARTpho, with varying diacritic removal rates. We examine specific predictions to estimate the models' ability to normalize diacritics in sentences and identify common errors encountered in these cases.

\begin{itemize}
    \item \textbf{Analysis of Example 1}: This sentence is entirely devoid of diacritics. In this scenario, no model is capable of completely translating the sentence accurately. However, as the removal rate $p$ increases, models tend to predict more correct words. ViSoBERT and BARTpho approach the closest accurate predictions at $p$ = 100\%.
    \item \textbf{Analysis of Example 2}: This sentence contains an abbreviation `vn' ($\text{`việt nam'}_\text{vietnam}$), and two words lacking diacritics `giai' ($\text{`giải'}_\text{tournament}$)) and `ơ' ($\text{`ở'}_\text{at}$). ViSoBERT fails to translate `vn', transforms `bay' (in $\text{`sân bay'}_\text{airport}$) unnecessarily into $\text{`bây'}_\text{you}$), and inaccurately handles diacritics by correctly adding them to `giải' but not to `ơ'. BARTpho, with $p$ = 50\%, provides the most accurate prediction. BARTpho also retains `bay' correctly without transforming it incorrectly as ViSoBERT does.
\end{itemize}

\subsubsection{Errors in Predictions of $D_U$ dataset}

We employ our proposed framework with BARTpho as Student model and $p$ = 100\%, to annotate the dataset $D_U$. As this dataset lacks ground truth labels, direct evaluation metrics cannot be applied. Instead, we will examine several examples to better understand the framework’s ability to normalize text when applied to unseen data.

\begin{table}[]
\centering
\begin{tabular}{ll}
\toprule
\textbf{ID}         & 1                                                                                                                                        \\
\textbf{Source}     & man hình rộng thoai mai xem phim.                                                                                                        \\
\textbf{English}    & Wide screen for comfortable watching movies.                                                                                             \\
\textbf{Prediction} & màn hình rộng thoải mái xem phim.                                                                                                        \\
\midrule
\textbf{ID}         & 2                                                                                                                                        \\
\textbf{Source}     & thuong wa di                                                                                                                             \\
\textbf{English}    & i feel so sorry                                                                                                                          \\
\textbf{Prediction} & thương quá đi                                                                                                                            \\
\midrule
\textbf{ID}         & 3                                                                                                                                        \\
\textbf{Source}     & \begin{tabular}[c]{@{}l@{}}bác @username cang ngày noi chuyện cang hay nge noi the nay\\ ung ho bác ne\end{tabular}                      \\
\textbf{English}    & \begin{tabular}[c]{@{}l@{}}@username, the more you talk, the more you hear things like this,\\ I support you\end{tabular}                \\
\textbf{Prediction} & \begin{tabular}[c]{@{}l@{}}bác @username càng ngày nói chuyện càng hay nghe nói thế này\\ ủng hộ bác nè\end{tabular}                     \\
\midrule
\textbf{ID}         & 4                                                                                                                                        \\
\textbf{Source}     & đôi vua chua sử dụng k biết ổn ap k nói chung là mua đc nhé m n                                                                          \\
\textbf{English}    & \begin{tabular}[c]{@{}l@{}}I haven't used it yet and don't know how whether it's ok. In general,\\ you can buy it, everyone\end{tabular} \\
\textbf{Prediction} & \begin{tabular}[c]{@{}l@{}}đôi vừa chưa sử dụng không biết ổn áp không nói chung là mua\\ được nhé mọi người\end{tabular}           \\
\bottomrule
\end{tabular}
\caption{Several High-Quality Predictions on $D_U$ dataset.}
\label{tab:error_goodPred}
\end{table}

Table \ref{tab:error_goodPred} presents examples where the framework effectively normalizes text from the $D_U$ dataset. Example 1 involves a sentence with multiple words lacking diacritics. The model accurately predicted and restored the diacritics for all the necessary words while leaving the already correct words unchanged. Example 2 features a sentence with both missing diacritics and teencode. Here, the model also achieved complete accuracy in its predictions. Example 3 is a sentence almost entirely devoid of diacritics, and the model successfully restored all the required diacritics. In the final example, the sentence contains a mix of words missing diacritics and abbreviations. The model provided generally accurate predictions, except for the initial word `đôi' (which should be `đội', meaning `wear' in English), where it failed to apply the correct normalization.

\begin{table}[]
\centering
\begin{tabular}{ll}
\toprule
\textbf{ID}         & 1                                                                                                                                                                                                              \\
\textbf{Source}     & \begin{tabular}[c]{@{}l@{}}shop tu vân nhiẻt tình dk tăng thêm túi đựng hạt chốh ầm,\\ giao hangg thiêu nhưg sau đo dk shop gui lại ngay!\end{tabular}                                                         \\
\textbf{English}    & \begin{tabular}[c]{@{}l@{}}The shop gave enthusiastic advice and was given an extra\\ moisture-proof bag of seeds, the delivery was missing,\\ but the shop sent it back immediately!\end{tabular}             \\
\textbf{Prediction} & \begin{tabular}[c]{@{}l@{}}shop từ vân nhiệtẻt tình được tăng thêm túi đựng hạt chốh\\ ầm, giao hàng thiêu nhưng sau đo được shop gửi lại ngay !\end{tabular}                                                  \\
\midrule
\textbf{ID}         & 2                                                                                                                                                                                                              \\
\textbf{Source}     & \begin{tabular}[c]{@{}l@{}}mình dùng thấy sp oki, pin bền cam nét, đa nghiệm nhanh, \\ nv thì nhiệt tình khỏi bàn luôn, hài lòng\end{tabular}                                                                  \\
\textbf{English}    & \begin{tabular}[c]{@{}l@{}}i found the product to be good, with long-lasting battery life,\\ sharp camera, fast multitasking, and very enthusiastic staff,\\ i’m satisfied.\end{tabular}                       \\
\textbf{Prediction} & \begin{tabular}[c]{@{}l@{}}mình dùng thấy sản o kí, pin bền cam nét, đa nghiệm nhanh\\ nhân vật thì nhiệt tình khỏi bàn luôn, hài lòng\end{tabular}                                                            \\
\midrule
\textbf{ID}         & 3                                                                                                                                                                                                              \\
\textbf{Source}     & máy muot cuc kì                                                                                                                                                                                                \\
\textbf{English}    & the machine works very smoothly                                                                                                                                                                                \\
\textbf{Prediction} & máy mượt cực kì kì                                                                                                                                                                                             \\
\midrule
\textbf{ID}         & 4                                                                                                                                                                                                              \\
\textbf{Source}     & \begin{tabular}[c]{@{}l@{}}tang ban 120.000 d de nap tien dien thoai viettel trong 06\\ thang ( 20.000 d / thang ) khi dang ky viettelpay ct danh\\ cho tb hoc sinh , sinh vien\end{tabular}                   \\
\textbf{English}    & \begin{tabular}[c]{@{}l@{}}receive a gift of 120,000 vnd for viettel mobile top-up\\ over 6 months (20,000 vnd/month) when you register for\\ viettelpay, this program is for student subscribers\end{tabular} \\
\textbf{Prediction} & \begin{tabular}[c]{@{}l@{}}tăng bạn 100 nghìn điểm để nạ tiền điện thoại viettel\\ trong 6 tháng ( 10.000 điểm / tháng ) khi đăngặng viettelpay\\ ct dành cho trung bình học sinh , sinh viên\end{tabular}    \\
\bottomrule
\end{tabular}
\caption{Several Low-Quality Predictions on $D_U$ dataset.}
\label{tab:error_badPred}
\end{table}

Table \ref{tab:error_badPred} presents examples where the framework not only cannot normalize text from the $D_U$ dataset but also make it become worsen:
\begin{itemize}
    \item \textbf{Example 1} is a sentence containing numerous errors: missing diacritics, abbreviations, and misspellings. The model was only able to correct a few missing diacritics (`gui lai' $\rightarrow$ $\text{`gửi lại'}_\text{send back}$) and accurately transcribe a few abbreviations and misspellings (`dk' $\rightarrow$ $\text{`được'}_\text{okay}$, `hangg' $\rightarrow$ $\text{`hàng'}_\text{goods}$). However, the model made several incorrect predictions (`nhiẻt' $\rightarrow$ `nhiệtẻt', `tu' $\rightarrow$ `từ' which should be `nhiệt' in $\text{`nhiệt tình'}_\text{enthusiastic}$ and `tư' in $\text{`tư vấn'}_\text{advise}$) or retained the original text when unable to make a prediction.
    \item \textbf{Example 2} contains relatively fewer errors, consisting only of two abbreviations: `sp' (for $\text{`sản phẩm'}_\text{product}$) and `nv' (for $\text{`nhân viên'}_\text{staff}$). However, the model failed to accurately predict these abbreviations (`sp' $\rightarrow$ `sản', `nv' $\rightarrow$ $\text{`nhân vật'}_\text{character}$). Additionally, the word $\text{`oki'}_\text{ok}$ in the original sentence did not require normalization but was incorrectly altered, resulting in a more inaccurate sentence.
    \item \textbf{Example 3} is a sentence entirely devoid of diacritics. While the model accurately predicted all words requiring diacritics, it added an unnecessary word `kì' (in $\text{`cực kì'}_\text{very}$) at the end of the sentence. 
    \item \textbf{Example 4} is an exceptionally long sentence, completely lacking diacritics and containing several abbreviations (`tb', `ct'). The model managed to normalize only a few words missing diacritics (`ban' $\rightarrow$ $\text{`bạn'}_\text{you}$, `tien dien thoai' $\rightarrow$ $\text{`tiền điện thoại'}_\text{telephone fee}$, `hoc sinh, sinh vien' $\rightarrow$ $\text{`học sinh, sinh viên'}_\text{student, undergraduate}$, etc.). In other cases, the model failed to provide correct translations. Furthermore, the model altered numerical values from correct to incorrect (`120.000 d' $\rightarrow$ `100 nghìn điểm', `20.000 d / tháng' $\rightarrow$ `10.000 điểm / tháng') and mistranslated many words (`nap' $\rightarrow$ `nạ' instead of $\text{`nạp'}_\text{recharge}$). Thus, in this instance, the model not only failed to accurately normalize the original sentence but also deteriorated its quality. 
\end{itemize}

\subsection{Evaluation on Downstream Tasks}

To evaluate the importance of lexical normalization, we conduct experiments on downstream tasks intergrated with a lexical normalization component. Specifically, we will apply our proposed framework to normalize raw input data for five downstream tasks: Hate Speech Detection, Emotion Recognition, Hate Speech Span Detection, Spam Review Detection, and Aspect-Based Sentiment Analysis. To assess the necessity of lexical normalization, we will use two key metrics: F1-micro and accuracy. These metrics provide a comprehensive view of the model's performance across different evaluation criteria.

\subsubsection{Hate Speech Detection}

Hate Speech Detection (HSD) is a significant challenge in NLP, aimed at identifying and classifying hate speech in text. Detecting and mitigating hate speech on social media platforms and online forums is crucial for ensuring a healthy and safe communication environment for users.

To address this problem, we refer to the ViHSD study \cite{10.1007/978-3-030-79457-6_35}, a research project focused on detecting hate speech in Vietnamese. In this study, two models - Text-CNN and GRU - were employed to predict and classify hate speech.

The team conducted experiments by applying their proposed framework to label the ViHSD dataset with canonical word forms and evaluated the performance of both the Text-CNN and GRU models before and after applying lexical normalization. The results were assessed using the F1-macro and Accuracy metrics.

\begin{table}[]
\centering
\begin{tabular}{ccccc}
\toprule
\textbf{Model}            & \textbf{Metric} & \textbf{Before} & \textbf{After} & \textbf{Improvement} \\
\midrule
\multirow{2}{*}{Text-CNN} & F1-macro (\%) & 58.17           & 60.74          & $\uparrow$2.57                 \\
                          & Accuracy (\%)   & 83.56           & 86.85          & $\uparrow$3.29                 \\
\midrule
\multirow{2}{*}{GRU}      & F1-macro (\%) & 58.41           & 60.31          & $\uparrow$1.90                 \\
                          & Accuracy (\%)   & 83.14           & 86.13          & $\uparrow$2.99                 \\
\bottomrule
\end{tabular}
\caption{Performance Comparison of Text-CNN and GRU Models on Hate Speech Detection Before and After Lexical Normalization.}
\label{tab:hsd}
\end{table}

\begin{figure}[ht]
    \centering
    \includegraphics[width=0.8\linewidth]{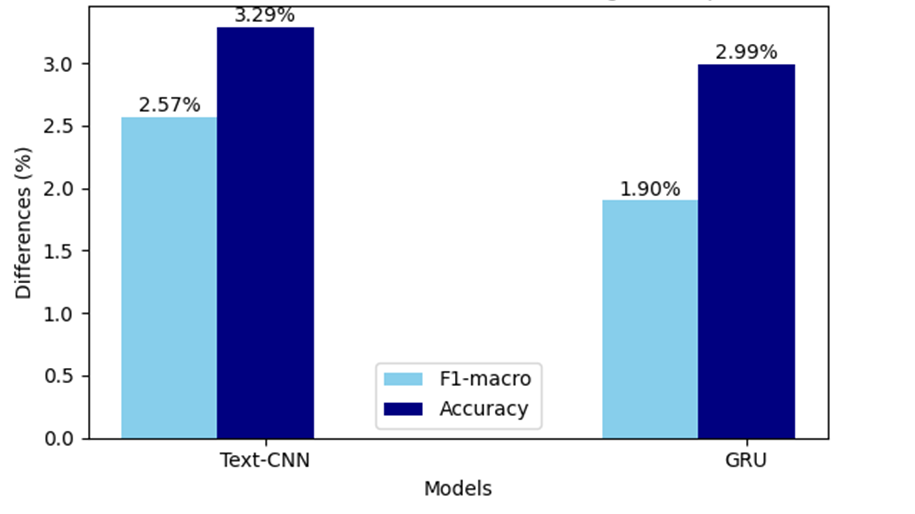}
    \caption{Metrics of Text-CNN and GRU Models on Hate Speech Detection Before and After Lexical Normalization.}
    \label{fig:hsd}
\end{figure}

As illustrated in Table \ref{tab:hsd} and Figure \ref{fig:hsd}, applying lexical normalization significantly enhanced the performance of Hate Speech Detection (HSD) in the Vietnamese context. Both the Text-CNN and GRU models showed marked improvements after normalization. Specifically, F1-macro score of Text-CNN increased by 2.57\%, and Accuracy rose by 3.29\%. The GRU model experienced a 1.90\% gain in F1-macro and a 2.99\% increase in Accuracy. While both models benefited from normalization, Text-CNN demonstrated greater overall improvement, suggesting it may better utilize normalized text features.

\subsubsection{Emotion Recognition}

Emotion Recognition (ER) involves identifying and classifying emotions within text. To explore this task, we refer to the study \cite{10.1007/978-981-15-6168-9_27}, which provides a comprehensive analysis of emotion recognition from Vietnamese text on social media platforms. This study employs two primary models for emotion prediction: Text-CNN and PhoBERT.

We conducted experiments using our proposed framework to normalize the UIT-VSMEC dataset and evaluated the performance of both models before and after lexical normalization. The results were assessed using F1-macro and Accuracy metrics.

\begin{table}[]
\centering
\begin{tabular}{ccccc}
\toprule
\textbf{Model}            & \textbf{Metric} & \textbf{Before} & \textbf{After} & \textbf{Improvement} \\
\midrule
\multirow{2}{*}{Text-CNN} & F1–macro (\%)   & 39.69           & 41.00          & $\uparrow$1.31                 \\
                          & Accuracy (\%)   & 49.35           & 51.37          & $\uparrow$2.02                 \\
\midrule
\multirow{2}{*}{PhoBERT}  & F1–macro (\%)   & 60.76           & 62.72          & $\uparrow$1.96                 \\
                          & Accuracy (\%)   & 63.05           & 63.06          & $\uparrow$0.01                 \\
\bottomrule
\end{tabular}
\caption{Performance Comparison of Text-CNN and GRU Models on Emotion Recognition Before and After Lexical Normalization.}
\label{tab:er}
\end{table}

\begin{figure}[ht]
    \centering
    \includegraphics[width=0.8\linewidth]{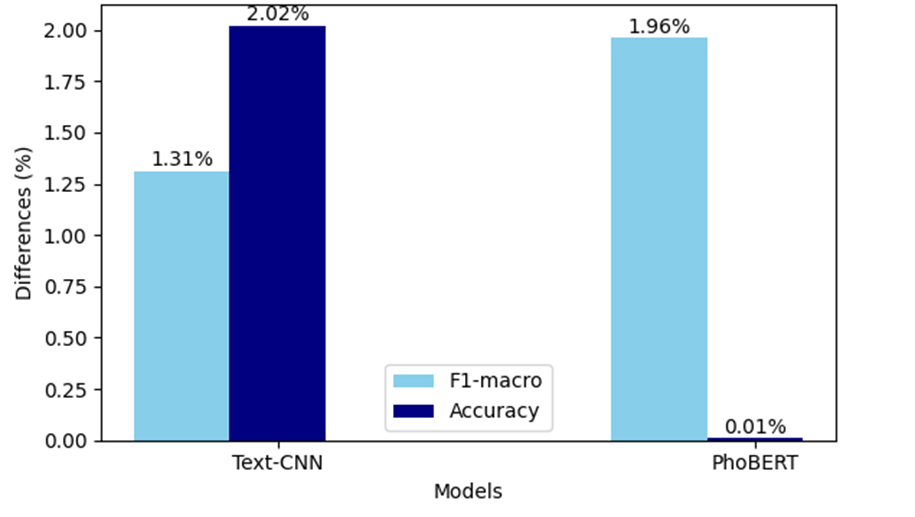}
    \caption{Metrics of Text-CNN and GRU Models on Emotion Recognition Before and After Lexical Normalization.}
    \label{fig:er}
\end{figure}

From the observations in Table \ref{tab:er} and Figure \ref{fig:er}, after applying lexical normalization, the Text-CNN model exhibited significant improvements in both F1-macro and Accuracy metrics. Notably, the greatest improvement was in Accuracy, with an increase of 2.02\%. Although the PhoBERT model showed only a marginal improvement in Accuracy (0.01\%), its F1-macro score increased by 1.96\%. This indicates that text normalization enhances the ability of PhoBERT model to classify emotions in less prevalent categories.

Both the Text-CNN and PhoBERT models showed improvements following lexical normalization, but the degree of improvement varied depending on the model. Text-CNN demonstrated a more substantial improvement in Accuracy, while PhoBERT showed significant enhancement in F1-macro. These findings highlight that lexical normalization is a crucial and beneficial step in improving the performance of emotion recognition models for Vietnamese text on social media.

\subsubsection{Hate Speech Span Detection}

Hate Speech Span Detection (HSSD) involves identifying and marking specific segments of text that contain hate speech within a larger document. In the ViHOS study \cite{hoang-etal-2023-vihos}, which provides an in-depth analysis of detecting hate speech spans in Vietnamese text, we utilized two primary models for prediction: XLM-R and PhoBERT.

We conducted experiments using our proposed framework on the ViHOS dataset and evaluated the performance of the XLM-R and PhoBERT models before and after applying lexical normalization. The results were assessed using F1-macro and Accuracy metrics.

\begin{table}[]
\centering
\begin{tabular}{ccccc}
\toprule
\textbf{Model}           & \textbf{Metric} & \textbf{Before} & \textbf{After} & \textbf{Improvement} \\
\midrule
\multirow{2}{*}{XLM-R}   & F1-macro (\%)   & 48.00           & 48.13          & $\uparrow$0.13                 \\
                         & Accuracy (\%)   & 86.71           & 86.73          & $\uparrow$0.02                 \\
\midrule
\multirow{2}{*}{PhoBERT} & F1-macro (\%)   & 47.21           & 47.49          & $\uparrow$0.28                 \\
                         & Accuracy (\%)   & 86.60           & 86.64          & $\uparrow$0.04                 \\
\bottomrule
\end{tabular}
\caption{Performance Comparison of Text-CNN and GRU Models on Hate Speech Span Detection Before and After Lexical Normalization.}
\label{tab:hssd}
\end{table}

\begin{figure}[ht]
    \centering
    \includegraphics[width=0.8\linewidth]{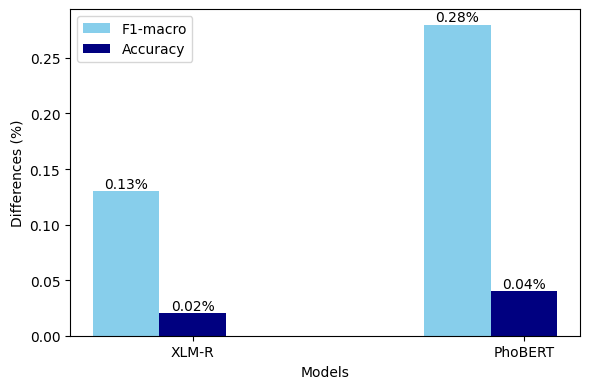}
    \caption{Metrics of Text-CNN and GRU Models on Hate Speech Span Detection Before and After Lexical Normalization.}
    \label{fig:hssd}
\end{figure}

As illustrated in Table \ref{tab:hssd} and Figure \ref{fig:hssd}, the application of lexical normalization yielded positive impacts on the performance of both the XLM-R and PhoBERT models in the task of Hate Speech Span Detection (HSSD). The result shows that while the improvements were modest, they were consistent across all metrics.

For the XLM-R model, the was an improvement of 0.13\% in F1-macro and 0.02\% in Accuracy. These increments, though small, suggest that lexical normalization provides a marginal but consistent benefit in enhancing the ability of model to correctly identify hate speech spans. The PhoBERT model exhibited a slightly higher improvement following normalization. Its F1-macro score witnessed an improvement of 0.28\%. Additionally, Accuracy increased from 86.60\% to 86.64\%, with a gain of 0.04\%. The greater relative improvement in F1-macro for PhoBERT compared to XLM-R indicates that PhoBERT may be more sensitive to the benefits of lexical normalization.

These findings highlight the value of text normalization in improving the performance of HSSD models. The consistent improvements across both models further affirm the role of preprocessing techniques like lexical normalization in refining the capabilities of emotion and hate speech detection systems.

\subsubsection{Spam Review Detection}

Spam Review Detection involves identifying and categorizing fake or fraudulent reviews within online product and service rating systems. This problem encompasses two primary tasks: detecting spam reviews (Spam Detection) and classifying the types of spam (Spam Classification).

In our study, we refer to the paper \cite{10.1007/978-3-031-21743-2_48}, which explores the detection and classification of spam reviews in Vietnamese. Various models were employed for prediction, including Text-CNN, GRU, and BiLSTM.

We conducted experiments using lexical normalization on the ViSpamReviews dataset and evaluated model performance before and after applying this normalization technique. Performance was measured using F1-macro and Accuracy metrics for both tasks: spam detection (Task 1) and spam classification (Task 2).

\begin{table}[]
\centering
\begin{tabular}{cccccc}
\toprule
\textbf{Model}            & \textbf{Task}      & \textbf{Metric (\%)} & \textbf{Before} & \textbf{After} & \textbf{Improvement} \\
\midrule
\multirow{4}{*}{Text-CNN} & \multirow{2}{*}{1} & F1-macro        & 77.04           & 77.28          & $\uparrow$0.24                 \\
                          &                    & Accuracy        & 83.31           & 83.67          & $\uparrow$0.36                 \\
                          & \multirow{2}{*}{2} & F1-macro        & 61.32           & 62.36          & $\uparrow$0.04                 \\
                          &                    & Accuracy        & 82.43           & 82.89          & $\uparrow$0.46                 \\
\midrule
\multirow{4}{*}{GRU}      & \multirow{2}{*}{1} & F1-macro        & 73.66           & 76.21          & $\uparrow$2.55                 \\
                          &                    & Accuracy        & 81.92           & 82.71          & $\uparrow$0.79                 \\
                          & \multirow{2}{*}{2} & F1-macro        & 62.72           & 62.94          & $\uparrow$0.22                 \\
                          &                    & Accuracy        & 81.79           & 81.67          & $\downarrow$0.12                 \\
\midrule
\multirow{4}{*}{BiLSTM}   & \multirow{2}{*}{1} & F1-macro        & 74.56           & 75.40          & $\uparrow$0.84                 \\
                          &                    & Accuracy        & 81.62           & 82.33          & $\uparrow$0.71                 \\
                          & \multirow{2}{*}{2} & F1-macro        & 55.19           & 62.13          & $\uparrow$6.94                 \\
                          &                    & Accuracy        & 81.01           & 81.85          & $\uparrow$0.84                 \\
\bottomrule
\end{tabular}
\caption{Performance Comparison of Text-CNN and GRU Models on Spam Review Detection Before and After Lexical Normalization.}
\label{tab:srd}
\end{table}

\begin{figure}[ht]
    \centering
    \includegraphics[width=\linewidth]{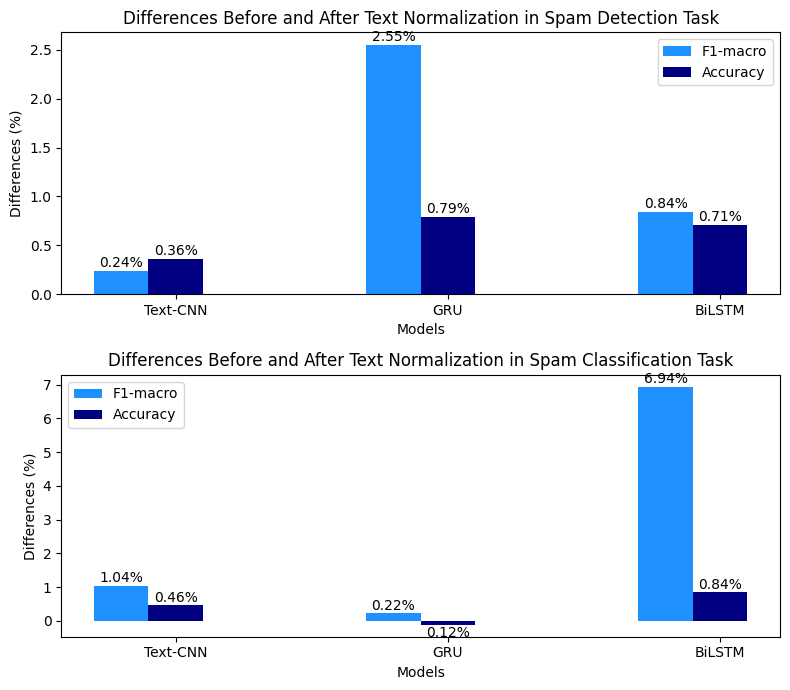}
    \caption{Metrics of Text-CNN and GRU Models on Spam Review Detection Before and After Lexical Normalization.}
    \label{fig:srd}
\end{figure}

From the results observed in Table \ref{tab:srd} and Figure \ref{fig:srd}, Text-CNN and BiLSTM models showed significant improvements across both tasks after applying lexical normalization. Specifically, Text-CNN saw a notable increase in F1-macro for both Task 1 (0.24\%) and Task 2 (0.04\%), alongside improvements in Accuracy. Similarly, BiLSTM demonstrated substantial enhancements, particularly in Task 2 where F1-macro rose significantly (from 55.19\% to 62.13\%), indicating improved performance in classifying different types of spam. GRU, while showing a substantial increase in F1-macro for Task 1 (2.55\%), experienced a slight decrease in Accuracy for Task 2 (0.12\%). This suggests that while GRU benefits from lexical normalization in detecting spam reviews more accurately, there may be nuances in its ability to classify them effectively.

Overall, all three models benefited from lexical normalization, highlighting its role in improving the models' ability to detect and classify spam reviews. Text-CNN and BiLSTM exhibited more consistent improvements across both tasks, underscoring their effectiveness in leveraging normalized text data for spam detection and classification.

These findings underscore the importance of preprocessing techniques like lexical normalization in enhancing the robustness and accuracy of models for spam review detection, and emphasize the need for careful consideration of model selection based on specific task requirements.

\subsubsection{Aspect-Based Sentiment Analysis}

Aspect-Based Sentiment Analysis (ABSA) is a task that involves detailed sentiment analysis in texts, focusing on specific aspects of the mentioned product or service. This task is typically divided into three main subtasks:

\begin{itemize}
    \item \textbf{Sentiment Classification}: Identifying the overall sentiment of the text (positive, negative, or neutral).
    \item \textbf{Aspect Detection}: Identifying specific aspects of the product or service mentioned in the text.
    \item \textbf{Aspect-Based Sentiment Detection}: Determining the sentiment related to each specific aspect.
\end{itemize}

We employed the PhoBERT model to perform all three tasks of ABSA. Our experiments involved applying lexical normalization to the ABSA dataset and evaluating the performance of PhoBERT model before and after normalization. We assessed results using F1-macro and Accuracy metrics across all tasks: Sentiment Detection, Aspect Detection, and Aspect-Based Sentiment Detection.

\begin{table}[]
\centering
\begin{tabular}{ccccc}
\toprule
\textbf{Task}                       & \textbf{Metric (\%)} & \textbf{Before} & \textbf{After} & \textbf{Improvement} \\
\midrule
\multirow{2}{*}{Sentiment}          & F1-macro           & 59.67           & 59.49          & $\downarrow$0.18                 \\
                                    & Accuracy             & 90.09           & 90.20          & $\uparrow$0.11                 \\
\midrule
\multirow{2}{*}{Aspect}             & F1-macro           & 59.60           & 60.74          & $\uparrow$1.14                 \\
                                    & Accuracy             & 90.90           & 90.95          & $\uparrow$0.05                 \\
\midrule
\multirow{2}{*}{Sentiment + Aspect} & F1-macro           & 51.10           & 51.61          & $\uparrow$0.51                 \\
                                    & Accuracy             & 90.09           & 90.20          & $\uparrow$0.11     
                                    \\
\bottomrule
\end{tabular}
\caption{Performance Comparison of Text-CNN and GRU Models on Aspect-Based Sentiment Analysis Before and After Lexical Normalization.}
\label{tab:absa}
\end{table}

Based on the results observed in the Table \ref{tab:absa}:

\begin{itemize}
    \item \textbf{Sentiment Classification}: Although there was a slight improvement in Accuracy (0.11\%), the F1-macro score decreased slightly (0.18\%). This suggests that while text normalization can improve overall performance, it may not yield uniform benefits across all sentiment classes.
    \item \textbf{Aspect Detection}: There was a more noticeable improvement in F1-macro (1.14\%), along with a slight improvement in Accuracy (0.05\%). This indicates that text normalization helps PhoBERT better identify specific aspects.
    \item \textbf{Aspect-Based Sentiment Detection}: All metrics showed slight improvements, particularly F1-macro (0.51\%), indicating that text normalization aids the model in better analyzing sentiment related to each aspect.
\end{itemize}

\begin{figure}[ht]
    \centering
    \includegraphics[width=0.9\linewidth]{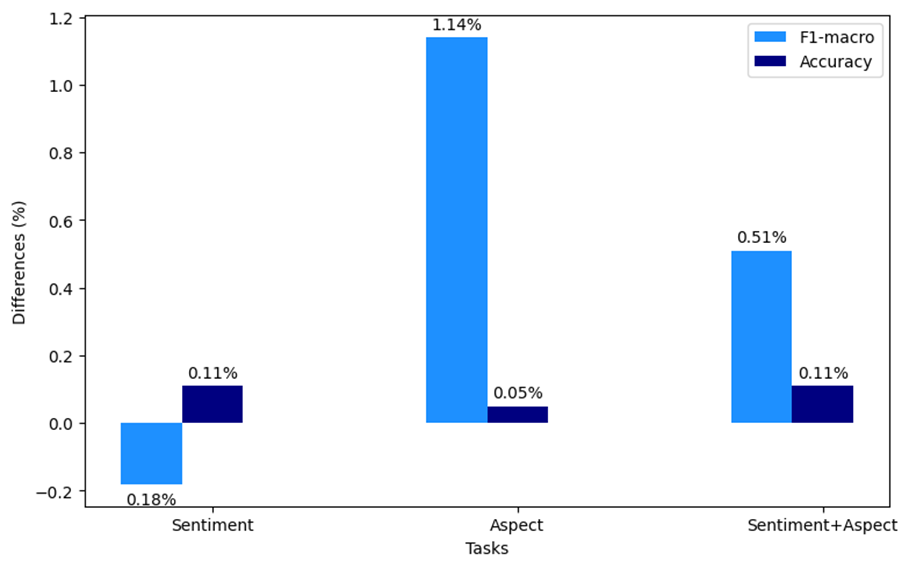}
    \caption{Metrics of Text-CNN and GRU Models on Aspect-Based Sentiment Analysis Before and After Lexical Normalization.}
    \label{fig:absa}
\end{figure}

According to Figure \ref{fig:absa}, PhoBERT demonstrated performance improvements in all ABSA tasks after applying lexical normalization, although the degree of improvement varied. Text normalization helps the PhoBERT model operate more effectively in identifying specific aspects and analyzing sentiment related to each aspect, especially as reflected in the F1-macro score.

These findings underscore the value of text normalization in enhancing the the efficiency of PhoBERT model for ABSA tasks, with notable gains in the ability of the model to detect and analyze sentiment related to specific aspects of the text.


\section{Conclusion and Future Works} \label{section-6}
This paper highlight the efficacy of our noval weakly supervised data labeling framework for lexical normalization, especially for low-resource languages like Vietnamese. This pioneering framework showcases robust performance across various base student models, including ViSoBERT, PhoBERT, and BARTpho. All model achieved lexical normalization accuracy exceeding 70\%, with BARTpho delivering the best results. When augmented with diacritic removal, BARTpho attained an impressive 85.64\% F1-score for NSWS, a 98.61\% correctness rate for non-normalized words, and a 96.16\% sentence-level accuracy.

The framework exhibited strong performance even on sentences not included in the training set, effectively normalizing common NSWs (such as $\text{`đc'}_\text{okay}$, $\text{`ko'}_\text{no}$) and abbreviations ($\text{`vn'}_\text{vietnam}$, $\text{`sv'}_\text{student}$). Furthermore, the implementation of lexical normalization significantly enhanced various downstream tasks. For instance, it improved accuracy and performance in Hate Speech Detection, Emotion Recognition, Hate Speech Span Detection, Spam Review Detection, and Aspect-Based Sentiment Analysis by 1-3\%. Among these, Hate Speech Detection saw the most notable improvement, with an increase of 3.29\% in accuracy and 2.57\% in the F1-macro score.

The lexical normalization capabilities of our framework on unlabeled datasets could not be directly evaluated due to the lack of ground truths. Training was resource-intensive without proportional performance gains, and the weak rules (regular expressions, dictionaries) used were not thoroughly optimized, limiting their effectiveness. The base model for lexical normalization needs refinement, especially in maintaining non-normalized words. We suggest several future directions as follows.

\begin{itemize}
    \item A comprehensive evaluation system will be developed to better assess the labeling capabilities of the weak supervision framework. If feasible, label the entire unlabeled dataset and use metrics for comparative analysis with accurate data. If direct labeling is impractical, explore alternative methods to evaluate dataset quality. Recent advancements in large language models could offer a promising approach for this assessment.
    \item Researchers can refine the weak rules within the framework (such as regular expressions and dictionaries) and incorporate additional rules. This enhancement will improve their supportive role and contribute to the overall accuracy of the labeling framework.
\end{itemize}

\section*{Declarations}

\textbf{Conflict of interest} The authors declare that they have no conflict of interest.

\section*{Funding}
This research was supported by The VNUHCM-University of Information Technology's Scientific Research Support Fund.

\section*{CRediT authorship contribution statement}
 \textbf{Dung Ha Nguyen:} Conceptualization; Formal analysis; Investigation; Methodology; Validation; Visualization; Writing - review \& original draft. \textbf{Anh Thi Hoang Nguyen:} Conceptualization; Data curation; Formal analysis; Investigation; Validation; Visualization; Writing - original draft. \textbf{Kiet Van Nguyen:} Conceptualization; Formal analysis; Investigation; Methodology; Validation; Supervision; Writing - review \& editing.

\bibliography{sn-bibliography}

\end{document}